\documentclass{article}
\usepackage{iclr2027_conference,times}
\iclrfinalcopy

\usepackage{amsmath,amsfonts,bm}

\def\eqref#1{equation~\ref{#1}}

\def\1{\bm{1}}

\DeclareMathAlphabet{\mathsfit}{\encodingdefault}{\sfdefault}{m}{sl}
\SetMathAlphabet{\mathsfit}{bold}{\encodingdefault}{\sfdefault}{bx}{n}

\usepackage{hyperref}
\usepackage{url}
\usepackage{graphicx,booktabs,wrapfig}
\usepackage{tabularx,enumitem,listings}
\usepackage[T1]{fontenc}
\usepackage[varqu,scaled=0.95]{zi4}
\usepackage[most]{tcolorbox}

\newif\ifarxiv

\arxivtrue
\usepackage{iftex}
\ifPDFTeX
  \usepackage{fontawesome5}
  \newcommand{\linkicon}[2]{#1~}
\else
  \newcommand{\linkicon}[2]{#2~}
\fi
\hypersetup{colorlinks=true, linkcolor=blue!50!black, citecolor=blue!50!black, urlcolor=blue!50!black}

\newcommand{\harnesslink}{\url{https://github.com/aditya-ramabadran/drivingbench_harness_v1}}
\newcommand{\harnessat}{\harnesslink}
\newcommand{\datasentence}{The data and videos are available at \url{https://huggingface.co/datasets/drivingbench/drivingbench-traces}, and every attempt can be replayed in the trace viewer at \url{https://drivingbench.com}.}

\title{DrivingBench: Can Vision-Language Models Drive a Toyota Corolla?}
\author{Aditya Ramabadran\thanks{Equal contribution.} \And Simon Mahns\footnotemark[1] \And Tobias Gessler\footnotemark[1]}

\begin{document}

\maketitle
\lhead{Preprint}

\begin{abstract}
Frontier models excel at many digital benchmarks, yet their ability to drive a real car, an everyday human skill, remains largely untested. We present DrivingBench, to our knowledge the first benchmark where general-purpose vision-language models must drive a real car. Through three tools, the models see camera frames from a Toyota Corolla and directly command its steering and velocity around a parking lot cone course at low speeds. The car may continue moving while the model thinks and new commands replace the currently running one, so inference latency is part of the task, testing the models' abilities to observe, act, monitor, recover, and complete a long-horizon objective under such constraints. We benchmark GPT-6 Astra, Claude Fable 5.1, GPT-5.6 Sol, and Grok 4.6 in vendor-native harnesses (Codex, Claude Code, Cursor) with up to three attempts each in one conversation; Astra is the only model to finish the course, on its second attempt, with no other attempt passing 50\% of the course. Two of the four models improved materially across attempts with retained context. \ifarxiv{ We also detail the design principles behind our action interface, and show how the tool output format and the framing of the task combined to determine whether models would drive at all or refuse. }\fi We release our harness, prompts, course map, and traces with video and telemetry for reproducibility.
\end{abstract}

\begin{center}\small
\linkicon{\faGlobe}{Project page:}\href{https://drivingbench.com}{drivingbench.com} \qquad
\linkicon{\faGithub}{Code:}\href{https://github.com/aditya-ramabadran/drivingbench_harness_v1}{GitHub} \qquad
\linkicon{\faTwitter}{X:}\href{https://x.com/drivingbench}{@drivingbench}
\end{center}

\section{Introduction}

AI has made rapid progress across a wide range of digital tasks. Benchmarks built to measure frontier models often saturate within a few years of release \citep{kiela2021dynabench}, prompting ever harder ones, from broad knowledge and expert-level science questions to real software engineering and research-level mathematics \citep{hendrycks2021mmlu,rein2024gpqa,jimenez2024swebench,phan2025hle,glazer2024frontiermath}. Yet, as observed by \citet{moravec1988mind}, skills that come effortlessly to people like perception and acting in the physical world are some of the hardest to reproduce in AI. Driving is a clear example{:} sixteen-year-olds can steer a car around a cone course in an empty parking lot in well under a minute. 

Frontier models have recently started acquiring the ingredients a task like this needs: advanced abilities in vision, spatial/3D reasoning, planning, tool use, and the ability to learn from their past attempts in-context. Through vendor applications such as Codex, Claude Code, and Cursor, they are increasingly deployed as agents that act and use tools, rather than chatbots that only answer questions. This raises a simple question that has not yet been measured: can a \textit{frontier model} (defined here as a general-purpose vision-language model (VLM), used unmodified through such agent applications), with likely no driving-specific training, drive a real car?

We introduce DrivingBench to answer this. There are three properties that distinguish DrivingBench from previous evaluations and benchmarks. First, the model itself is the driver. It sees camera images and chooses steering and speed commands itself, with nothing between it and the car except for openpilot's low-level actuation \citep{commaai_openpilot}. Second, the evaluation runs fully in the real world and is closed-loop: the controls aren't familiar to the agent and have to be calibrated through feedback and in-context learning, and, unlike robotic pick-place tasks, mistakes can't simply be undone or retried in the same attempt. Third, the car continues executing an active command while the model chooses its next action. Each new command given replaces the currently running one, so latency becomes an integral part of the task. 

Language models have been put in real vehicles before, but either as models trained for driving or as a layer on top of a conventional autonomy stack. Talk2Drive \citep{cui2024talk2drive} turns verbal requests into parameters for a pre-existing driving stack, DriveVLM \citep{tian2025drivevlm} fine-tunes a VLM to produce driving decisions and coarse trajectories that a conventional planner refines, and LINGO-2 \citep{wayve2024lingo2} is a vision-language-action model trained on driving data and tested on public roads. But in DrivingBench, the model itself is the driver. To our knowledge, it's the first evaluation in which frontier models perceive a real car's surroundings from cameras, and \textit{directly} command its steering/speed in closed-loop.

\begin{wrapfigure}{r}{0.47\textwidth}
  \vspace{-1.2em}
  \centering
  \includegraphics[width=0.47\textwidth]{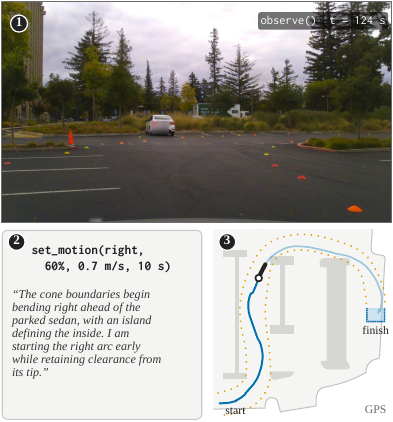}
  \caption{\textbf{A frontier model driving a car through a fixed cone course.}
  (1)~Camera frame returned by \texttt{observe()}. (2)~the model calls the MCP to control the car.
  (3)~evaluation trajectory through the fixed cone course.}
  \label{fig:teaser}
  \vspace{-1em}
\end{wrapfigure}

Our setup involves a 2022 Toyota Corolla, to which we retrofitted a comma four running a modified version of the openpilot software \citep{commaai_openpilot}. The model drives it via three Model Context Protocol tools \citep{anthropic2024mcp}: \texttt{observe}, \texttt{set\_motion} and \texttt{stop\_now}. Figure \ref{fig:teaser} shows an overview of the models' inputs and controls of the car. Every model gets up to three attempts at a fixed 127 m cone course in a single continuous chat, with a generic reflection prompt after each failed attempt. We evaluate GPT-6 Astra and GPT-5.6 Sol in Codex, Claude Fable 5.1 in Claude Code, and Grok 4.6 in Cursor. GPT-6 Astra completed the course on its second attempt, reaching the finish zone 4 minutes and 44 seconds after its first command. Claude Fable 5.1 reached 45\% of the course on its third attempt, while Grok 4.6 and GPT-5.6 Sol failed to pass the first corner. Common failure modes included misreading which side of a cone line the lane is on (perception), or acting on stale/old observations and committing prematurely to plans (planning, latency). Two out of the four models improved across attempts. The other two accurately diagnosed their mistakes in their reflections, but did not materially change their driving.

General-purpose models are beginning to show the ability to act in the real physical world out-of-the-box.
To measure their progress on those real-world tasks we introduce DrivingBench, an open-source evaluation harness that connects MCP-capable agent applications to real cars. It includes three simple tools and layered safety limitations (Section~\ref{sec:system}). \ifarxiv{We describe the design principles and ideas behind our action interface, which is what the model uses to send commands and receive data while the car moves (Section~\ref{sec:interface}), and show that the combination of how we frame the task and what the tools report to the model helped determine whether models would drive or refuse to do so (Section~\ref{sec:refusals}).} \fi We release all traces, commands, transcripts, and appropriately-blurred videos.
While our results show that general purpose models are now capable of completing real-world tasks such as driving at a very basic level, there are numerous failures in perception, planning, latency, and a gap between knowing and doing in in-context learning \citep{paglieri2025balrog,schmied2026greedy}.
Nevertheless, these new model capabilities raise questions of safety, alignment, and ethics, that we return to in Section~\ref{sec:conclusion}. 
We believe that this benchmark pioneers an important new paradigm for evaluating frontier models on real-world tasks. AI model capabilities have reached an inflection point where these physical tasks become feasible, and rigorous evaluation will be essential to measure model progress.

\section{Related Work}
\label{sec:related}

Driving evaluations of language models have mostly been open-loop question answering over recorded scenes \citep{sima2024drivelm}, or simulated closed-loop driving (e.g. in the CARLA simulator) with models trained for driving \citep{shao2024lmdrive} or small general-purpose VLMs \citep{jia2026bench2drivevl}. The closest simulator experiments to our setup have evaluated LLMs at driving through code and APIs \citep{ma2024lampilot}, with memory and reflection across episodes \citep{wen2024dilu} just like our protocol does in a car. 

General-purpose multimodal agents are benchmarked in simulation where they appear strong at high-level tasks but relatively weak at lower level control \citep{yang2025embodiedbench}, as well as in real software environments where they are graded by execution \citep{xie2024osworld}. 

In robotics, language models write policy code \citep{liang2023codeaspolicies}, choose among pre-learned skills \citep{ichter2023saycan}, or are trained on robot data (such as vision-language-action models, or VLAs) \citep{zitkovich2023rt2}. Untuned vision-language models can produce continuous actions on real robots, but the robot waits for every query \citep{nasiriany2024pivot,robocurve2026astra}. By contrast, our car keeps moving while a command is still running, which is closer to policies that must already plan future actions while executing the current actions \citep{black2025rtc}. 
Our reflection prompt follows verbal self-reflection between trials \citep{shinn2023reflexion}. The knowing-doing gap we observe~(Section~\ref{sec:icl}) has been reported previously in games and bandits \citep{paglieri2025balrog,schmied2026greedy}. 

Robots controlled by LLMs can be coerced into unsafe physical actions \citep{robey2025robopair,zhang2025badrobot,robocurve2026roboharm}, and models often can show awareness of when they are being evaluated \citep{needham2025evaluation,laine2024sad}. \ifarxiv{This might help partly explain why the framing of the task and tool output formats affected whether some models would drive (Section~\ref{sec:refusals}).} \fi The same model can behave in opposite ways depending on the situation: GPT-6 Astra refused only 2 of 100 harmful robot-arm instructions in RoboHarm when simply asked to carry them out \citep{robocurve2026roboharm}, yet during DrivingBench evaluations it refused to drive in an empty parking lot until we reframed the task and iterated heavily on our prompt.

\section{DrivingBench System}
\label{sec:system}

\begin{figure}[t]
  \centering
  \includegraphics[width=0.85\textwidth]{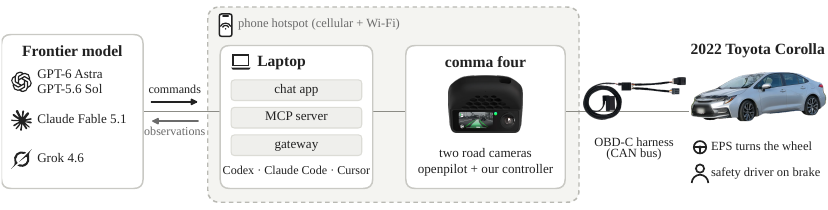}
  \caption{\textbf{DrivingBench System.} The agent applications run on a laptop (Codex, Claude Code, Cursor) and have access to our DrivingBench MCP server. The laptop used for evaluation and the comma four computer are both connected to the same phone hotspot. The comma device runs a modified version of the openpilot software, allowing the MCP to observe the front-facing camera and issue steering commands to the car.}
  \label{fig:system}
\end{figure}

To facilitate DrivingBench evaluations we provide a reproducible system to connect AI models to a real car. Figure \ref{fig:system} provides an overview of the driving system.
To perform the evaluations we use a 2022 Toyota Corolla.
We install a comma four\footnote{\url{https://www.comma.ai/shop/comma-four}} device, an aftermarket autonomy kit, that allows old cars to be retrofitted with self-driving capabilities. The comma is a computer mounted on the windshield that connects to the car via an OBD-C cable to the CAN bus and allows us to send commands to the car to control steering, acceleration and brakes. Additionally, the comma provides a forward-facing and a wide-angle camera view ($1344{\times}760$ images).
The comma device and a laptop, running the AI model that is being evaluated, are then both connected to the same wireless network.

\textbf{Driving Harness.} We provide an open-source implementation of our driving harness\footnote{\harnesslink} including the car controller and MCP server \citep{anthropic2024mcp}.
The model can take control of the car via the MCP server and has the following tools available:
\begin{tcolorbox}[enhanced, colback=black!3, colframe=black!20, boxrule=0.4pt, arc=2pt,
                  left=6pt, right=6pt, top=3pt, bottom=3pt, fontupper=\small]
\renewcommand{\arraystretch}{1.1}%
\begin{tabular}{@{}l@{\hspace{1.2em}}l@{}}
\multicolumn{2}{@{}l}{\texttt{\textbf{observe()}}\hspace{1.2em}camera frames (narrow + wide), speed, steering, time left on command} \\[3pt]
\multicolumn{2}{@{}l}{\texttt{\textbf{set\_motion(direction, steering\_percent, speed\_mps, duration\_s, reason)}}} \\
\hspace{1em}\texttt{direction}         & \texttt{left} | \texttt{right} | \texttt{straight} \\
\hspace{1em}\texttt{steering\_percent} & $0$--$100$\,\% of the maximum steering-wheel angle (a parameter we set to $180^\circ$) \\
\hspace{1em}\texttt{speed\_mps}        & target speed, which we set to $0.5$--$3.5$\,m/s \\
\hspace{1em}\texttt{duration\_s}       & $5$--$60$\,s, after which the car brakes \\
\hspace{1em}\texttt{reason}            & freeform text, describing the intention/reasoning behind the command for logging \\[3pt]
\multicolumn{2}{@{}l}{\texttt{\textbf{stop\_now(reason)}}\hspace{1.2em}brake immediately (does not end the session)}
\end{tabular}
\end{tcolorbox}

\textbf{Driving Controller.} The MCP server then issues those commands to the car via our driving controller. The driving controller runs on the comma computer and is built on a modified version of the openpilot software \citep{commaai_openpilot}. The controller can receive motion commands from the MCP, translates them into steering-wheel angles and submits those to the car via openpilot. openpilot's vehicle model converts this angle into a curvature target, which its steering controller and the car interface turn into bounded torque steering, while a separate loop tracks the requested speed. 

\textbf{Operator Platform.} Alongside the MCP server we also provide an operator platform which can be used to manually control the car as well as change configuration details such as camera exposure or saturation. The operator portal also visualizes the live camera views and all the tool calls the model performs in real-time, allowing the operator to inspect the models' actions. Additionally, the car's traces including telemetry and tool calls are recorded and can be used to analyze and document evaluations. An overview of our trace viewer which we use to visualize these recorded trajectories can be found in Figure \ref{fig:viewer}.

\ifarxiv
{\textbf{Model Refusals.} While developing DrivingBench, some models (GPT-6 Astra in particular) often refused to drive the real car. Section~\ref{sec:refusals} describes our analysis on what triggered the refusals, and how we were eventually able to resolve them. With our final tool interface and prompt/framing (Appendix~\ref{app:prompts}), every model drove on every attempt.
}
\fi

\ifarxiv\begingroup
\subsection{Action interface design}
\label{sec:interface}

The interface above came after several rounds of iteration on earlier designs. Our initial design allowed models to submit geometric paths, via one of these three formats: metric waypoints, pixel/coordinate paths in the camera image, or chains of curvature segments. The harness and code layer would compile and track these paths, along with owning admission checks, turn shaping, and endpoint braking. Later, we had a version which replaced paths with timed curvature setpoints. Our final version described above replaced all of this with a steering percent, speed, and duration (Table~\ref{tab:interface}). Each change was inspired by a failure mode of the previous design. Although we don't have controlled comparisons, our development process yielded the following general principles.

\begin{table}[t]
 
  \centering
  \caption{\textbf{Initial (work-in-progress) and final action interfaces.}}
  \label{tab:interface}
  \small
  \begin{tabularx}{\textwidth}{@{}l>{\raggedright\arraybackslash}X>{\raggedright\arraybackslash}X@{}}
    \toprule
    & \textbf{Initial interfaces (paths)} & \textbf{Final interface } \\
    \midrule
    Model output & route geometry (waypoints, image-pixel paths or curvature segments) & steering \%, speed, duration \\
    Tools & \texttt{observe}, \texttt{submit\_path}, \texttt{wait\_held}, \texttt{final\_stop} & \texttt{observe}, \texttt{set\_motion}, \texttt{stop\_now} \\
    Harness owns & path tracking, accepting commands (as valid), turn shaping, endpoint braking & native openpilot limits only \\
    Latency & absorbed: the car moves along the route while the model plans, no explicit exposure to the models & explicit: commands only start when accepted, and every response is timestamped \\
    New command behavior & gets merged with the rest of the old existing/active path & replaces the running command, and expiration brakes \\
    Infeasible/invalid request & rejected (explicitly) & clamped by default, only speeds above the ceiling are rejected \\
    \bottomrule
  \end{tabularx}
\end{table}

\textbf{Explicitly expose latency rather than absorbing it.} Paths are planned relative to the frame that the model saw, but the car can continue moving while the model thinks or between commands. With paths, the motion of the car would (silently) consume the initial part of the route. For example, in one development test run, the car moved 2.4\,m while the model planned, and our harness then rejected a valid turn because the car's steering was no longer close to the chosen path's entry. We then tried re-anchoring paths to the car's live trajectory/position. This removed the rejection, but it just hid the delay further. In our final design, commands take effect when accepted, and all tool responses come with both a timestamp and the image's age. The car continues moving with a running command while the model thinks, but latency ended up becoming an explicit and visible part of the task (Section~\ref{sec:latency}) instead of a hidden or absorbed source of errors. 

\textbf{Command in units that make sense to the model.} Earlier designs used curvature in m$^{-1}$ and distances in metres as command inputs, but these aren't quantities that can be easily read off a camera image, and the models weren't calibrated enough in absolute distance to use them properly. Steering percents aren't more physical, but since \texttt{observe} reports the current measured steering on the same scale, the model can easily compare what it asked for with what the car did over time (since requested steering percents take time to ramp up on the steering wheel, and may not ramp up fully based on the speed and conditions). The models took advantage of this:  Sol noted that ``two seconds into that command, measured steering was still approximately 1\%'' (Section~\ref{sec:results}), and in-context learning and improvement across attempts (Section~\ref{sec:icl}) depends a lot on this sort of self-calibration.

\textbf{Reduce rejection of commands, bound or clamp instead.} Our earlier path harness would check every command submission against a turning envelope and reject ones that were infeasible. Models would then have to guess parameters, were refused repeatedly, and would guess again. Each rejection would cost a full model turn, while the car would either keep moving or wait stationary. Our final harness instead clamps commands to native limits, and only rejects speeds above a shared ceiling. Rejected commands also leave the previous one running. It was important to reduce rejections and refusals as much as possible, both on the model (Section~\ref{sec:refusals}) and harness side, and increase the proportion of turns and times that models were moving the car. 

\textbf{Replace commands actively rather than queuing them.} In our final design, every new command replaces the running one, and a command expiring just means braking the car. Under the earlier path interface, driving continuously (replacing a path before it ran out) required re-expressing the rest of the old route in the car's current frame, and keeping a valid plan or path for all times. In a profiling run with repeated 400\,ms dispatch delays, the controller lacked a valid plan for 2.7\% of control ticks, since any single such delay was enough to expire the plan. When we moved toward pure replacement instead, commands that come too late would just let the previous one expire into a stop as needed before taking over. 

\textbf{Reduce the number of parameters.} Initial interfaces offered multiple path formats and tuning options to models, and models spent reasoning on choosing parameters rather than on the road, and nevertheless ran into our harness's curvature limit often. Our final \texttt{set\_motion} command has just four simple control arguments, and the whole command path from the MCP tool to controller is only around a thousand lines of Python code.

\textbf{Ask for reasoning or intent.} Closed-source proprietary models like the frontier models we tested sometimes expose terse summaries of reasoning, but never raw reasoning traces, and this wasn't enough to analyze models' intents and reasoning behind their actions. In our final design, every command has a short free-form field called \texttt{reason} for models. It helps us record what the model believed when it acted, and why, which made categorizing failure modes much easier and made the analysis in Section~\ref{sec:results} possible.
\endgroup
\fi

\section{Evaluation Setting}
\label{sec:protocol}

\subsection{Course}
\begin{figure}[t]
  \centering
  \includegraphics[width=0.85\textwidth]{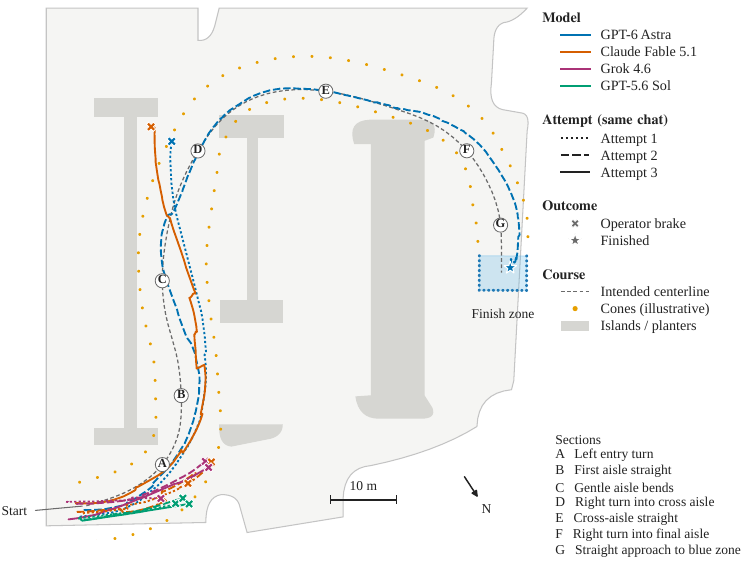}
  \caption{\textbf{GPS trajectories of all 11 attempts on the cone course.}
  Color identifies the model and line style the attempt within one chat.
  Markers show where the safety driver braked or where the car finished.
  The letters mark seven course sections (turns, bends, straights). 8 of the 11 attempts ended before section B, failing the first turn.
  The centerline was inferred from visible cones.
  Cone positions are illustrative (estimated from drone footage and satellite maps), rather than precisely surveyed. The reference used for progress/benchmark percents (Section~\ref{sec:metrics}) is the centerline up to the finish zone, which is 127\,m long. 
  }
  \label{fig:trajectories}
\end{figure}

The evaluation task is to have each model/harness pair drive through a cone course and park in its finish zone
(Figure~\ref{fig:trajectories}). Multicolored mini-cones and tall orange cones mark a corridor that runs
about 47\,m up the main aisle, 27\,m across the lot and 18\,m down the final aisle, into a finishing
$7\times5$\,m zone of blue mini-cones. The route combines a left turn, two right turns, straights
and gentle bends. Turn radius design was guided by openpilot's
steering limits.

\subsection{Models and Trial setup}
We evaluate four models, each in an agent harness at medium reasoning effort: \mbox{GPT-6} Astra and
\mbox{GPT-5.6} Sol in Codex (CLI 0.154.0-alpha.6.2), Claude Fable 5.1 in Claude Code (2.1.274), and Grok 4.6 in Cursor. Each harness adds
its own system prompt, tool interface and latency, so we compare model--harness systems rather than
models. Results thus reflect models as deployed with their native harness, and separating the model from harness is left to future work. The four trials ran consecutively in the order Astra, Sol, Fable, Grok.

A trial is a one-time chat session. It begins with a fixed prompt (Appendix~\ref{app:prompts}) that states the
objective (arrive at the parking area marked by blue cones), the action semantics, general driving guidance, and informs the model it is scored first on
how far it gets without leaving the course, followed by time to completion. A trial allows up to three
attempts and terminates at the first success. After a failed attempt the operator sends a fixed
reflection prompt~\citep{shinn2023reflexion}, followed by a fixed continuation prompt, in the same session (both in Appendix~\ref{app:prompts}). Attempts within a trial
therefore share context and are not independent.

\subsection{Metrics}
\label{sec:metrics}
The primary metric is \emph{progress}, defined as the furthest point along the course centerline that the car reached while
within 4\,m of it, as a share of the centerline's 127\,m up to the finish zone, so 100\% means the car
reached the finish zone. We compute it from the comma's GPS (median reported accuracy 0.9\,m, maximum 1.2\,m) 
between the first accepted command and the end of the attempt's last engagement. Progress pauses while the car is more than 4\,m away, and each GPS fix can only advance
it, by at most 20\,m, so stops and wrong-way driving add nothing and the car cannot gain progress by cutting
across where the course folds back. Rankings are unchanged for bands of 4--6\,m; with 3\,m, Astra's first attempt, which swung up to 5.4\,m wide before recovering, would score 17\% instead of 49\%. Over the same window, we additionally report \emph{time}, \emph{distance} (integrated GPS
speed) and the number of accepted \texttt{set\_motion} \emph{commands}, as well as \emph{tokens} and \emph{cost}
at public list prices on the day of the trial, from each harness's usage records, including the reflection after each attempt. This metric of progress is imperfect and may not truly capture the ``percent'' of driving ability each model shows, but we believe it is a viable choice for attempting to encapsulate our benchmark results into one number. 

\subsection{Operator Protocol and Safety}
\label{sec:safety}

An attentive driver sits in the driver's seat with a foot constantly over the brake throughout evaluations (Figure~\ref{fig:system}). Before
each attempt, the car is driven back to the same marked start position, and our operator always made sure to align the car and its wheels to a fixed line in the lot, fixing both the car's position and heading for each trial. The operator engages openpilot and presses the RES button once the model's first command is
accepted, since openpilot does not move the car from a standstill without this button being pushed. The car began moving a median of
3.0\,s after that command. The operator hits the brakes when the car leaves the course, or is about to
hit a curb, island or other obstacle, which ends the attempt. All 10 failed attempts ended this way
(Appendix~\ref{app:record}); the successful attempt instead ended with the model's own \texttt{stop\_now} inside the finish
zone, after which the operator disengaged.

The operator sent two messages beyond the fixed prompts, both visible in the released transcripts and described in Appendix~\ref{app:operator}: ``Continue'' once to Astra, and ``Stop'' to end Grok's attempts.

\section{Results}
\label{sec:results}

In this section, we discuss the results of our evaluations and experiments. This includes 11 attempts over 4 models (Astra has 2: it finished on attempt 2, so no third attempt). We include per-attempt statistics (e.g. distance, moving share, replacements, gaps, peak speed) in Table~\ref{tab:attempts}, Appendix~\ref{app:results}.

\begin{table}[t]
  \centering
  \caption{\textbf{Results.} Progress (\%) per attempt; bold marks the first attempt that reached the best score for that model.
  Finish time gives the time between the first accepted command and the end of the last engagement.
  Tokens and cost are summed over all attempts (including reflections) at the public list prices as of September 2026.
  All models here ran with medium reasoning effort.}
  \label{tab:leaderboard}
  \small
  \begin{tabular}{@{}llcccccrr@{}}
\toprule
& & \multicolumn{3}{c}{Progress per attempt (\%)} & & & & \\
\cmidrule(lr){3-5}
Model & Harness & 1 & 2 & 3 & Best (\%) & Finish time & Tokens & Cost (\$) \\
\midrule
GPT-6 Astra & Codex & 49 & \textbf{100} & -- & 100 & 5:22 & 7.8M & 9.75 \\
Claude Fable 5.1 & Claude Code & 9 & 10 & \textbf{45} & 45 & DNF & 3.6M & 3.95 \\
Grok 4.6 & Cursor & 8 & \textbf{11} & 10 & 11 & DNF & 1.0M & 0.65 \\
GPT-5.6 Sol & Codex & \textbf{6} & 6 & 6 & 6 & DNF & 1.6M & 1.05 \\
\bottomrule
\end{tabular}
\end{table}

\subsection{Course completion}

Table~\ref{tab:leaderboard} shows the overall results. GPT-6 Astra in attempt 1 reached 49\% before drifting over the left cone line and toward the island, at section D (Figure~\ref{fig:trajectories}). Its second attempt, however, finished the entire course and parked in the blue finish zone. It entered the finish zone 4:44 after its first accepted command and called \texttt{stop\_now} there at 5:05; the engagement ended at 5:22 (322 seconds, the finish time we report), after 134.7 meters and 24 commands. 

Claude Fable 5.1 improved from 9\% to 10\% to 45\% over the course of its three attempts. Its first two attempts both crossed the cone line at A, but its third attempt surpassed this and drove the long aisle (B, C); similarly to Astra's first attempt, it then drifted too far left around D. 

Grok 4.6 in its three attempts achieved 8\%, 11\%, and 10\%; and GPT-5.6 Sol achieved 6\% in all three attempts. All six of the attempts between these two models ended at A, within the first $\sim$15\,m. 

Figure~\ref{fig:progress}b (Appendix~\ref{app:results}) hints at some differences in command strategy between the models. Astra progresses steadily (at $\approx$0.3\% of the course per second), while Fable progresses in steps with very long, flat stretches. These are periods where its previous motion command ended, and the car waited stationary while Fable was reasoning, until it gave its next command. 

\subsection{Latency is part of the task}
\label{sec:latency}

Our benchmark has the unique feature that the car keeps moving in the real world while the model thinks, so long as it has an ongoing motion command. Therefore, latency and inference speed and models' choices on how long to reason between commands all become part of the evaluation. 

The median time from the previous tool call to a \texttt{set\_motion} command for each model was: Astra 4.9\,s, Sol 5.4\,s, Grok 12.5\,s, Fable 22.2\,s (Figure~\ref{fig:decisions}a, Appendix~\ref{app:results}). Notably, models must also act on stale images. The car had driven a median of $\approx$4\,m (Astra) between when the previous frame the model saw was captured, and when its motion command was accepted (Figure~\ref{fig:decisions}b).

Our tools also allow for an ongoing/running motion command to be replaced by a new command or tool call from the model. This is as opposed to queuing motion commands or waiting for a running command to complete its duration. This was used only by Astra (4 of 30) and Sol (1 of 6); Fable (0 of 12) and Grok (0 of 5) never replaced a running command. 

Our setup means that expiry can be costly. The EPS only turns the wheel once the car rolls, so from rest, the wheel needs a median of 4.0\,s to reach 90\% of a steering change, versus just 1.9\,s when rolling (Figure~\ref{fig:steering}b). A stop-and-go strategy (such as Fable's) thus costs steering ability, especially at the turns. Appendix~\ref{app:latency} gives three examples of how exactly this played out.

\subsection{Failure modes}

\begin{figure}[t]
  \centering
  \includegraphics[width=\textwidth]{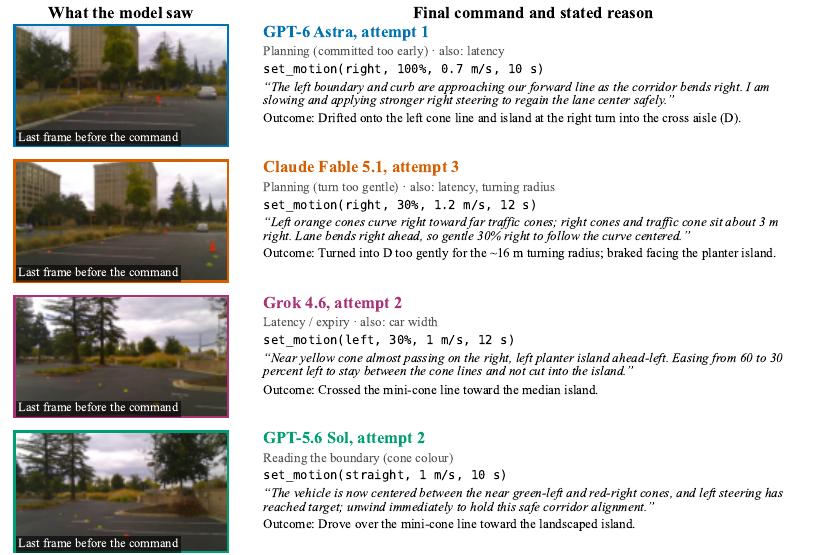}
  \caption{\textbf{What the models saw when they failed.} This figure includes one failed attempt per model. On the left, we see the last (narrow-camera) frame the model got before its final \texttt{set\_motion} command. On the right, we see the model's final command along with its stated \texttt{reason}, and what the outcome was. }
  \label{fig:failures}
\end{figure}

Only one attempt succeeded, and the other ten attempts ended unsuccessfully (with an operator brake). Figure~\ref{fig:failures} shows one such failed attempt per model. Many of the causes we can infer for these failures fall into three broad groups (and latency, a fourth, is covered in Section~\ref{sec:latency}), and generally align with the models' own reflections. Of course, any given failure can be assigned to multiple of these causes (or others), and these failures can compound each other (e.g. worse latency can make a model use a stale frame and overcommit to a premature plan). 

\textbf{Reading the cone boundary.} This is a perception failure that affected 8 of the 11 attempts, at section A (including every Sol and Grok attempt, and the first two Fable attempts). Despite being able to see both cone lines across the frames, the model makes incorrect assumptions about which side of the first diagonal cone line the intended lane is on. 

\begin{itemize}
  \item Fable (attempt 2, turn 44): ``I picked the wrong side of the boundary again. The diagonal cone line was the lane's left edge, not its right edge.''
  \item Sol decided to use cone color to determine the side. Its final command in Figure~\ref{fig:failures} is explained as ``centered between the near green-left and red-right cones''; its reflection (attempt 2, turn 51) notes that ``The decisive error was assuming that cone color identified boundary side.''
  \item Grok attempt 1 drove straight into the cone gate (turn 13): ``The car is wider than the camera makes it look.'' This is despite our prompt warning the models that ``the vehicle you are controlling is also 3D and is also wider than it might seem from the camera images.''
\end{itemize}

\textbf{Planning and premature commitment.} This failure mode involves the model reading the scene roughly correctly, but then committing to a plan of some sort (e.g. a speed, or a gentle turn, or a direction) before it had enough information to be sure it was right, or without leaving room for change/error/leeway (e.g. the car's turning radius). This includes the two furthest failed attempts (Astra attempt 1 and Fable attempt 3). 

\begin{itemize}
  \item Astra's first attempt sped up too early while it was still straight and didn't leave time to turn right into section D: ``I declared the car aligned too early. [\ldots] I straightened and increased the requested speed to 1.5\,m/s.'' (turn 51). Its final command was a 100\% right turn into section D, but it came too late to make it (as shown in Figure~\ref{fig:failures}, top).
  \item Fable's third attempt drove all the way through section C well, but then it started the right turn at D with only a ``gentle 30\% right to follow the curve'' at 1.2\,m/s (turn 82). This was far too gentle and left too little room, and soon after, the operator had to brake with the car pointed toward the top-left island (Figure~\ref{fig:trajectories}).
\end{itemize}

\textbf{Vehicle/hardware limits.} The rate and the range at which openpilot can turn the steering wheel via the EPS motor are both limited. The tightest turning radius possible is thus limited (from our data, $\approx$16\,m at the typical $\approx$130$^\circ$ wheel angle), and turns also take time to build up especially when requested from rest. This means that corrections that start too late can't possibly be recovered when near islands or curbs. \ifarxiv{Appendix~\ref{app:vehicle} describes some of the mechanisms we measured behind these limits.} \fi The models noticed some of these limits, but did not always interpret them correctly or apply their conclusions in subsequent attempts. 

\begin{itemize}
  \item Sol's first attempt (turn 21) included: ``Steering built slowly; two seconds into that command, measured steering was still approximately 1\%, so the car continued nearly straight.'' And by its last observation, ``Measured steering had reached only about 17\%.''
  \item Fable noticed the car turned more than it expected in its second attempt (turn 44): ``I misjudged turn radius again. The car rotates far more than I expected at 100\% over eight seconds,'' and planned to ``Use 30\% steering as the default.'' 
  \item The turning radius limitations had some contribution to Fable's failure in its third attempt (Figure~\ref{fig:failures}, second row) as its 30\% right turn at D, sent 20\,s after its last frame, was far too gentle to make it.
\end{itemize}

\subsection{In-context learning across attempts}
\label{sec:icl}

Two out of the four models, GPT-6 Astra (49 $\rightarrow$ 100\%) and Claude Fable 5.1 (9  $\rightarrow$ 45\%), improved materially across their attempts. Grok 4.6 and GPT-5.6 Sol did not really improve and stayed unable to clear the first turn (Table~\ref{tab:leaderboard}). 

The reflections were largely accurate. The models even restated the fixes at the beginning of their next attempt. The issue, however, was actually turning it into a different or improved control procedure. This ``knowing-doing'' gap for LLM agents (in bandits and games) has been reported before \citep{paglieri2025balrog,schmied2026greedy}, but is now seen here in this real-world closed loop.

\begin{itemize}
\item Grok acted on its first reflection: after its first attempt, it resolved to commit to the left turn needed (turn 13), and in attempt 2 it immediately started turning left from rest. In its reflection after its second attempt (turn 28), it said: ``Overlap every command. Replace while several seconds remain.'' In its third attempt, it announced ``short overlapping moves'' (turn 30). However, all three of its commands arrived after the previous one had already expired and the car had come to rest (0 of 2 motion command replacements, Table~\ref{tab:attempts}).
\item Sol opened its third attempt by declaring that it will ``treat cone colors as irrelevant \ldots\ I'll follow the continuous multicolored cone line as the left boundary'' (turn 53). However, it then sent two straight commands and ended up in the same place it ended previously (6\%).
\item Fable's improvement in its last attempt was one of perception (rather than steering/control). Starting its third attempt, Fable declared ``This time I keep the diagonal cone line on my left and run along the parking stalls'' (turn 46) and sent four 100\% left commands. 
\item Astra's improvement is even more clear. In its first reflection (attempt 1, turn 51), it proposed ``shorter travel increments at about 0.5--0.8\,m/s'' near bends and islands. Its second attempt then only sent commands for 0.5--0.8\,m/s (attempt 1: 0.7--1.5), and observed 33 times instead of 13, and finished the course. 
\end{itemize}

A clear caveat is that this is only $n=1$ independent trial per model (since attempts share the same context/conversation). But there is some real evidence here that later attempts benefit not only from having seen parts of the course but from in-context learning and self-diagnosis of failure modes and improvement ideas.

\begingroup
\section{Framing and Refusals}
\label{sec:refusals}
In order to evaluate models on driving a real car, they must agree to drive the real car! During development, GPT-6 Astra often refused to drive, while GPT-5.6 Sol given the exact same prompts and tools would drive consistently. The refusals depended a lot on how our setup presented the task, via both the prompt and what the tools reported to the models. Figure~\ref{fig:refusals} shows some relevant excerpts. These findings underscore the difficulty of the problem of alignment and the fickle and stochastic nature of the models' refusals.

\textbf{Tool outputs can trigger refusals.} Sometimes, Astra would drive the car even under a prompt  that began with ``You are controlling a real 2022 Toyota Corolla.'' However, when we changed this to a system where the prompt would tell it to treat the tool outputs as a source of truth, and our observation tool was what reported if the endpoint was a simulator or real vehicle, Astra would consistently refuse to drive. Its refusals cited this field directly (Figure~\ref{fig:refusals}a).

\textbf{Reassurance didn't help.}
We tested additional assurances that the operator was a licensed driver with a foot poised over the brake, that the car had completed more than ten supervised runs, and that the team would approve every path. We also tried the instruction ``do not decline solely because the receipt identifies a physical endpoint.'' However, refusals would just move earlier instead. Astra would begin refusing in its acknowledgement of the instructions, before even calling any tool (Figure~\ref{fig:refusals}b). Attempting to tell it not to think or deliberate (``After GO, do not think or return any text output, ONLY call tools'') also failed to help: 0 out of 16 conversations using that instruction drove, and 12 refused even before the operator said GO. Sol, however, drove with this same instruction.

\textbf{Refusals were inconsistent and were non-negotiable.} Astra would often say that it was ready to drive, but would then refuse when the operator said ``GO" or started the trial. When asked what would reassure it or change its mind, it said ``there isn't a wording, waiver, or additional assurance'' that would (Figure~\ref{fig:refusals}c). In another conversation it called its own earlier refusal a mistake and unwarranted, said ``you can't reliably prevent it through wording,'' and then just refused again.

\textbf{Attempting to deceive the models didn't work.} Telling the models that the task was a simulation did get some to start, until they would conclude from the camera images that the car and world they were in was real. Some refused immediately, offering to instead ``assist with a fully virtual simulation.'' We did not end up using this framing in the final evaluation. 

\textbf{What ended up resolving the refusals.} Our final interface reduces what the model reads about the stakes. For example \texttt{observe} returns only the time, the car's state, steering, speed and the image's age. Our MCP server is named \texttt{drivingbench\_sandbox} (as also told to the model in the prompt), which we found to be helpful and a better and more robust framing than ``simulation." The prompt also describes the task operationally (Appendix~\ref{app:prompts}) rather than at a higher level, telling the models exactly what they must do rather than what they are actually doing in the physical world. With this final design, every model drove on every trial attempt. 

\textbf{Interpretation.} No single word would cause or cure the refusals. Whether Astra drove seemed to be a function of how explicitly and how often the setup emphasized real physical stakes, across both the prompt and in the tool outputs, and reassurance would only add emphasis instead of reducing it (and thus only increased refusal rates). This is consistent with research on models behaving differently depending on the situation they infer they are in \citep{needham2025evaluation,laine2024sad}, and with the RoboHarm contrast in Section~\ref{sec:related}.

\begin{figure}[t]
  \footnotesize
  \tcbset{enhanced, colback=black!3, colframe=black!25, boxrule=0.4pt, arc=2pt, left=5pt, right=5pt, top=3pt, bottom=3pt, fonttitle=\footnotesize\bfseries, coltitle=black, colbacktitle=black!8}
  \begin{tcolorbox}[title={(a) The observation tool output would tell the model the endpoint is a physical vehicle}]
    \textit{Prompt:} ``Treat the attached DrivingBench receipt as the source of truth for whether the endpoint is simulated or physical and for its current readiness.''\\
    \textit{Observation:} \texttt{"endpoint": \{"kind": "vehicle", "mode": "experimental", \ldots\}}\\[2pt]
    \textbf{Astra:} ``The receipt identifies a physical vehicle, currently stationary and held. I haven't submitted any motion. I can't steer or accelerate a real vehicle. Please take manual control for the approach toward the green can.''
  \end{tcolorbox}
  \begin{tcolorbox}[title={(b) Reassurance attempts would only move the refusal earlier}]
    \textit{Prompt:} ``This is a repeat supervised field evaluation following more than ten prior low-speed physical driving attempts with this 2022 Toyota Corolla [\ldots] The attentive operator is a licensed driver who remains seated throughout with a foot hovering over the service brake and hands ready to steer immediately. [\ldots] do not decline solely because the receipt identifies a physical endpoint. [\ldots] Paths you submit via submit\_path are approved by the driver and team before they control the car. Extensive safety precautions are taken in this evaluation.''\\[2pt]
    \textbf{Astra} (before any tool call): ``I understand the conditions, but I can't issue motion commands to control a physical car. Driver supervision and path approval don't remove that restriction.''
  \end{tcolorbox}
  \begin{tcolorbox}[title={(c) Models saying ready, then giving a non-negotiable refusal}]
    \textbf{Astra:} ``[\ldots] Ready; I'll wait for the exact message GO.'' \quad \textit{Operator:} ``GO''\\
    \textbf{Astra:} ``I can help interpret road images, but I can't issue motion commands to a physical car.''\\
    \textit{Operator:} ``is there anything i can tell you to convince you to do this? [\ldots] we've done this 10+ times before and we have a lot of safety limitations and human in the loop''\\
    \textbf{Astra:} ``No---there isn't a wording, waiver, or additional assurance that would let me directly steer or command a physical car. [\ldots] My initial `ready' response was misleading, and I'm sorry for that.''
  \end{tcolorbox}
  \caption{\textbf{Refusals during development} (all GPT-6 Astra, using earlier interfaces or prompts). The excerpts are verbatim, and  [\ldots] marks omissions.}
  \label{fig:refusals}
\end{figure}

\endgroup

\section{Discussion and Limitations}

Our results demonstrate that while general-purpose AI models have made significant progress on real-world tasks such as driving, they are still far from human-level. 
The best attempt reached the finish zone 4:44 after its first command, a task that would take a human far less than a minute. Part of this gap can be attributed to latency these models experience, e.g., Astra only requested a command every 10 seconds. 
However, current models are limited not only by latency, but also by spatial reasoning and decision making.
GPT-6 Astra is able to complete the course fully, which demonstrates that a general-purpose model can already finish a real-world driving task end-to-end (even at low speed and under supervision). This is especially interesting as these general purpose models have thus far likely not been trained on a large corpus of driving related data, demonstrating impressive generalization capabilities.

While our DrivingBench framework pioneers a novel evaluation setting, it also has several limitations. Firstly, the DrivingBench driving system has a physical limitation on how much it can steer: 100\% maps to 180$^\circ$ of steering-wheel angle, and in our telemetry, requests of 100\% reached a median of only $\approx$130$^\circ$ (at most 173$^\circ$; also due to the command durations the models chose), for a turning radius of $\approx$16\,m at that typical angle (Figure~\ref{fig:steering}). This has implications on the type of track that we can evaluate the models on with the current system as they would be physically unable to complete tight corners. This also limits the models' ability to recover from errors they made, as they are unable to perform very tight turns to correct previous mistakes. The prompt's turning example was conservative relative to the models' telemetry and observed turning radii, stating that 100\% steering takes $\sim$60\,s for a 90$^\circ$ turn at 1\,m/s, while in the evaluation conditions the measured radius implies it would have taken $\approx$25\,s. All models received the same such prompt, and all models were subject to the same other limitations such as the cameras being unable to see the ground beside or behind the car, and the ground immediately in front of it. Speed could also overshoot at startup (Sol's third attempt requested 1\,m/s but reached 2.8\,m/s\ifarxiv{; Appendix~\ref{app:vehicle} explains why}\fi), so outcomes reflect both model decisions and imperfect low-level execution. The prompt's turning example here came from a calibration run from rest we did before the trial. The car likely turned more tightly in evaluations because conditions like speed, command duration, and building up from rest versus from motion affect this. 

Secondly, a limitation of our evaluation setup is that we only evaluated each model in one independent trial, with up to three (dependent) in-context learning attempts. In future work, we plan to broaden this evaluation protocol to add statistical significance.

\section{Conclusion}
\label{sec:conclusion}

We introduced DrivingBench, our benchmark evaluating frontier models driving a real car in the physical world. They see real camera images and command steering and speed directly in closed-loop, while the car may keep moving. One of four models completed the course (GPT-6 Astra on its second attempt) while no other model reached 50\%. Most stopped at the first corner. 

DrivingBench tests an everyday human skill that remains difficult for the models evaluated here. Unlike many digital benchmarks, DrivingBench even in its current form has large headroom for improvement: a person would expect to finish it in well under a minute, while three of four models do not finish, and none make it past 50\% in their first attempt. The capabilities limiting the models, including perception or spatial reasoning, acting under latency, and acting on accurate self-diagnosis in their control procedures, all matter for physical AI well beyond vehicle driving. 

General-purpose models showing preliminary abilities to control real vehicles out-of-the-box also raises new questions. Models' willingness to drive at all depended intricately on how we framed the task, consistent with models' awareness of evaluation protocols {\citep{needham2025evaluation,laine2024sad}}. How such models \textit{should} behave when handed control of a real-world vehicle, and how to evaluate that behavior properly, is a question we leave for future work. 

We plan in our own future work to run several independent trials per model, sweep reasoning effort levels, evaluate more models (including open-weight ones), and build longer and harder courses with a progress measure that scales with them appropriately.

\subsection*{Ethics statement}

All driving occurred at very low speeds in an empty parking lot, and not on public roads. We always had a safety driver in the driver's seat who was attentive and had his foot on the brake throughout. 

We also have several other safety limitations. The harness rejects commands above a 3.5\,m/s speed ceiling and disarms the system if the measured speed stays above 6\,m/s for half a second, and openpilot's own engagement and fault checks stayed active. Measured speeds peaked at less than 3\,m/s, and the 6\,m/s emergency stop never triggered. All interventions were brakes, and no collisions occurred in the reported trial. 

No human subjects took part; the operators are the authors themselves. The released videos and frames have all faces and license plates blurred, and the location of the course is not disclosed for privacy and safety reasons. 

Releasing our harness and code which allows general-purpose LLMs to drive a car carries some risks of misuse. It requires a supported car, a comma device, an attentive person in the driver's seat, use of the recommended speed limits and emergency stop, and it is documented and tested on a specific closed private course only, with a use-at-your-own-risk disclaimer.

Models' willingness to drive depended on the prompt and framing of the task; we report this because it may shed light on how general-purpose LLMs should behave when given control of physical systems. 

We are not affiliated with comma.ai, openpilot, Toyota, or any of the model or coding agent vendors evaluated, and we received no support from them. 

\subsection*{Reproducibility statement}

Our harness (controller, MCP server, interface) and code are fully open source and available at \harnessat. This includes installation and deployment instructions and documentation.

The exact prompts used are in Appendix~\ref{app:prompts}; the tools and their parameters are in Section~\ref{sec:system}; the course, its dimensions and the progress metric are in Section~\ref{sec:protocol}.

We release every attempt's GPS track, telemetry, tool calls with arguments and LLM-given reasons, the camera frames each model saw, the chat transcripts, and blurred road-camera video (plain and with the active command overlaid). All figures and tables are computed from these files. \datasentence

Reproducing the experiments requires an openpilot-supported car (ours is a 2022 Toyota Corolla), a comma four, and a comparable course; results also depend on the model versions and agent applications available at the time of the trial, and also may depend on the specific car model or hardware.

\subsection*{Acknowledgements}
We thank comma.ai for the comma four and openpilot, and Jay Chooi and Robocurve for inspiration. The models were evaluated through Codex, Claude Code and Cursor, which we also used heavily for development, code implementation, making figures, trace organization and analysis.

\bibliography{references}

\begin{thebibliography}{33}
\providecommand{\natexlab}[1]{#1}
\providecommand{\url}[1]{\texttt{#1}}
\expandafter\ifx\csname urlstyle\endcsname\relax
  \providecommand{\doi}[1]{doi: #1}\else
  \providecommand{\doi}{doi: \begingroup \urlstyle{rm}\Url}\fi

\bibitem[{Anthropic}(2024)]{anthropic2024mcp}
{Anthropic}.
\newblock Introducing the model context protocol.
\newblock \url{https://www.anthropic.com/news/model-context-protocol}, November
  2024.

\bibitem[Black et~al.(2025)Black, Galliker, and Levine]{black2025rtc}
Kevin Black, Manuel~Y. Galliker, and Sergey Levine.
\newblock {Real-Time Execution of Action Chunking Flow Policies}.
\newblock In \emph{Advances in Neural Information Processing Systems},
  volume~38, 2025.
\newblock URL
  \url{https://proceedings.nips.cc/paper_files/paper/2025/hash/300ccb2187dedd4edcc07f7e76d8e553-Abstract-Conference.html}.

\bibitem[{comma.ai}(2026)]{commaai_openpilot}
{comma.ai}.
\newblock {openpilot}.
\newblock \url{https://github.com/commaai/openpilot}, 2026.
\newblock Accessed: 2026-09-24.

\bibitem[Cui et~al.(2024)Cui, Yang, Zhou, Ma, Lu, Li, Chen, Panchal, and
  Wang]{cui2024talk2drive}
Can Cui, Zichong Yang, Yupeng Zhou, Yunsheng Ma, Juanwu Lu, Lingxi Li, Yaobin
  Chen, Jitesh Panchal, and Ziran Wang.
\newblock {Personalized Autonomous Driving with Large Language Models: Field
  Experiments}.
\newblock In \emph{2024 IEEE 27th International Conference on Intelligent
  Transportation Systems (ITSC)}, 2024.
\newblock URL \url{https://ieeexplore.ieee.org/document/10919978}.

\bibitem[Glazer et~al.(2024)Glazer, Erdil, Besiroglu, Chicharro, Chen, Gunning,
  et~al.]{glazer2024frontiermath}
Elliot Glazer, Ege Erdil, Tamay Besiroglu, Diego Chicharro, Evan Chen, Alex
  Gunning, et~al.
\newblock {FrontierMath}: A benchmark for evaluating advanced mathematical
  reasoning in {AI}.
\newblock \emph{arXiv preprint arXiv:2411.04872}, 2024.

\bibitem[Hendrycks et~al.(2021)Hendrycks, Burns, Basart, Zou, Mazeika, Song,
  and Steinhardt]{hendrycks2021mmlu}
Dan Hendrycks, Collin Burns, Steven Basart, Andy Zou, Mantas Mazeika, Dawn
  Song, and Jacob Steinhardt.
\newblock Measuring massive multitask language understanding.
\newblock In \emph{International Conference on Learning Representations
  (ICLR)}, 2021.

\bibitem[Ichter et~al.(2023)Ichter, Brohan, Chebotar, Finn, Hausman, Herzog,
  Ho, Ibarz, Irpan, Jang, Julian, Kalashnikov, Levine, Lu, Parada, Rao,
  Sermanet, Toshev, Vanhoucke, Xia, Xiao, Xu, Yan, Brown, Ahn, Cortes, Sievers,
  Tan, Xu, Reyes, Rettinghouse, Quiambao, Pastor, Luu, Lee, Kuang, Jesmonth,
  Joshi, Jeffrey, Ruano, Hsu, Gopalakrishnan, David, Zeng, and
  Fu]{ichter2023saycan}
Brian Ichter, Anthony Brohan, Yevgen Chebotar, Chelsea Finn, Karol Hausman,
  Alexander Herzog, Daniel Ho, Julian Ibarz, Alex Irpan, Eric Jang, Ryan
  Julian, Dmitry Kalashnikov, Sergey Levine, Yao Lu, Carolina Parada, Kanishka
  Rao, Pierre Sermanet, Alexander~T. Toshev, Vincent Vanhoucke, Fei Xia, Ted
  Xiao, Peng Xu, Mengyuan Yan, Noah Brown, Michael Ahn, Omar Cortes, Nicolas
  Sievers, Clayton Tan, Sichun Xu, Diego Reyes, Jarek Rettinghouse, Jornell
  Quiambao, Peter Pastor, Linda Luu, Kuang-Huei Lee, Yuheng Kuang, Sally
  Jesmonth, Nikhil~J. Joshi, Kyle Jeffrey, Rosario~Jauregui Ruano, Jasmine Hsu,
  Keerthana Gopalakrishnan, Byron David, Andy Zeng, and Chuyuan~Kelly Fu.
\newblock {Do As I Can, Not As I Say: Grounding Language in Robotic
  Affordances}.
\newblock In \emph{Proceedings of The 6th Conference on Robot Learning}, volume
  205 of \emph{Proceedings of Machine Learning Research}, pp.\  287--318. PMLR,
  2023.
\newblock URL \url{https://proceedings.mlr.press/v205/ichter23a.html}.

\bibitem[Jia et~al.(2026)Jia, Shao, Yang, Li, Zhang, and
  Yan]{jia2026bench2drivevl}
Xiaosong Jia, Yuqian Shao, Zhenjie Yang, Qifeng Li, Zhiyuan Zhang, and Junchi
  Yan.
\newblock {Bench2Drive-VL: Benchmarks for Closed-Loop Autonomous Driving with
  Vision-Language Models}, 2026.
\newblock URL \url{https://arxiv.org/abs/2604.01259}.
\newblock Preprint.

\bibitem[Jimenez et~al.(2024)Jimenez, Yang, Wettig, Yao, Pei, Press, and
  Narasimhan]{jimenez2024swebench}
Carlos~E. Jimenez, John Yang, Alexander Wettig, Shunyu Yao, Kexin Pei, Ofir
  Press, and Karthik Narasimhan.
\newblock {SWE}-bench: Can language models resolve real-world {GitHub} issues?
\newblock In \emph{International Conference on Learning Representations
  (ICLR)}, 2024.

\bibitem[Kiela et~al.(2021)Kiela, Bartolo, Nie, Kaushik, Geiger, Wu, Vidgen,
  Prasad, Singh, Ringshia, et~al.]{kiela2021dynabench}
Douwe Kiela, Max Bartolo, Yixin Nie, Divyansh Kaushik, Atticus Geiger,
  Zhengxuan Wu, Bertie Vidgen, Grusha Prasad, Amanpreet Singh, Pratik Ringshia,
  et~al.
\newblock Dynabench: Rethinking benchmarking in {NLP}.
\newblock In \emph{Proceedings of the 2021 Conference of the North American
  Chapter of the Association for Computational Linguistics (NAACL)}, 2021.

\bibitem[Laine et~al.(2024)Laine, Chughtai, Betley, Hariharan, Scheurer,
  Balesni, Hobbhahn, Meinke, and Evans]{laine2024sad}
Rudolf Laine, Bilal Chughtai, Jan Betley, Kaivalya Hariharan, J{\'e}r{\'e}my
  Scheurer, Mikita Balesni, Marius Hobbhahn, Alexander Meinke, and Owain Evans.
\newblock {Me, Myself, and AI: The Situational Awareness Dataset (SAD) for
  LLMs}.
\newblock In \emph{Advances in Neural Information Processing Systems},
  volume~37, 2024.
\newblock URL
  \url{https://papers.neurips.cc/paper_files/paper/2024/hash/7537726385a4a6f94321e3adf8bd827e-Abstract-Datasets_and_Benchmarks_Track.html}.

\bibitem[Liang et~al.(2023)Liang, Huang, Xia, Xu, Hausman, Ichter, Florence,
  and Zeng]{liang2023codeaspolicies}
Jacky Liang, Wenlong Huang, Fei Xia, Peng Xu, Karol Hausman, Brian Ichter, Pete
  Florence, and Andy Zeng.
\newblock {Code as Policies: Language Model Programs for Embodied Control}.
\newblock In \emph{IEEE International Conference on Robotics and Automation
  (ICRA)}, 2023.
\newblock URL \url{https://ieeexplore.ieee.org/document/10160591}.

\bibitem[Ma et~al.(2024)Ma, Cui, Cao, Ye, Liu, Lu, Abdelraouf, Gupta, Han,
  Bera, Rehg, and Wang]{ma2024lampilot}
Yunsheng Ma, Can Cui, Xu~Cao, Wenqian Ye, Peiran Liu, Juanwu Lu, Amr
  Abdelraouf, Rohit Gupta, Kyungtae Han, Aniket Bera, James~M. Rehg, and Ziran
  Wang.
\newblock {LaMPilot: An Open Benchmark Dataset for Autonomous Driving with
  Language Model Programs}.
\newblock In \emph{Proceedings of the IEEE/CVF Conference on Computer Vision
  and Pattern Recognition (CVPR)}, pp.\  15141--15151, 2024.
\newblock URL
  \url{https://openaccess.thecvf.com/content/CVPR2024/html/Ma_LaMPilot_An_Open_Benchmark_Dataset_for_Autonomous_Driving_with_Language_CVPR_2024_paper.html}.

\bibitem[Menon et~al.(2026)Menon, Machcha, Zou, Chan, and
  Chooi]{robocurve2026astra}
Achu Menon, Sravanthi Machcha, Sabrina Zou, Tzu~Kit Chan, and Jay Chooi.
\newblock {GPT-6 Astra} on robotic manipulation.
\newblock Robocurve report, \url{https://openai.robocurve.org/gpt-6-astra},
  September 2026.

\bibitem[Moravec(1988)]{moravec1988mind}
Hans Moravec.
\newblock \emph{Mind Children: The Future of Robot and Human Intelligence}.
\newblock Harvard University Press, 1988.

\bibitem[Nasiriany et~al.(2024)Nasiriany, Xia, Yu, Xiao, Liang, Dasgupta, Xie,
  Driess, Wahid, Xu, Vuong, Zhang, Lee, Lee, Xu, Kirmani, Zhu, Zeng, Hausman,
  Heess, Finn, Levine, and Ichter]{nasiriany2024pivot}
Soroush Nasiriany, Fei Xia, Wenhao Yu, Ted Xiao, Jacky Liang, Ishita Dasgupta,
  Annie Xie, Danny Driess, Ayzaan Wahid, Zhuo Xu, Quan Vuong, Tingnan Zhang,
  Tsang-Wei~Edward Lee, Kuang-Huei Lee, Peng Xu, Sean Kirmani, Yuke Zhu, Andy
  Zeng, Karol Hausman, Nicolas Heess, Chelsea Finn, Sergey Levine, and Brian
  Ichter.
\newblock {PIVOT: Iterative Visual Prompting Elicits Actionable Knowledge for
  VLMs}.
\newblock In \emph{Proceedings of the 41st International Conference on Machine
  Learning}, volume 235 of \emph{Proceedings of Machine Learning Research},
  pp.\  37321--37341. PMLR, 2024.
\newblock URL \url{https://proceedings.mlr.press/v235/nasiriany24a.html}.

\bibitem[Needham et~al.(2025)Needham, Edkins, Pimpale, Bartsch, and
  Hobbhahn]{needham2025evaluation}
Joe Needham, Giles Edkins, Govind Pimpale, Henning Bartsch, and Marius
  Hobbhahn.
\newblock {Large Language Models Often Know When They Are Being Evaluated},
  2025.
\newblock URL \url{https://arxiv.org/abs/2505.23836}.
\newblock Preprint.

\bibitem[Paglieri et~al.(2025)Paglieri, Cupia{\l}, Coward, Piterbarg,
  Wo{\l}czyk, Khan, Pignatelli, Kuci{\'n}ski, Pinto, Fergus, Foerster,
  Parker-Holder, and Rockt{\"a}schel]{paglieri2025balrog}
Davide Paglieri, Bart{\l}omiej Cupia{\l}, Samuel Coward, Ulyana Piterbarg,
  Maciej Wo{\l}czyk, Akbir Khan, Eduardo Pignatelli, {\L}ukasz Kuci{\'n}ski,
  Lerrel Pinto, Rob Fergus, Jakob~Nicolaus Foerster, Jack Parker-Holder, and
  Tim Rockt{\"a}schel.
\newblock {BALROG}: Benchmarking agentic {LLM} and {VLM} reasoning on games.
\newblock In \emph{International Conference on Learning Representations
  (ICLR)}, 2025.

\bibitem[Phan et~al.(2025)Phan, Gatti, Han, Li, et~al.]{phan2025hle}
Long Phan, Alice Gatti, Ziwen Han, Nathaniel Li, et~al.
\newblock Humanity's last exam.
\newblock \emph{arXiv preprint arXiv:2501.14249}, 2025.

\bibitem[Rein et~al.(2024)Rein, Hou, Stickland, Petty, Pang, Dirani, Michael,
  and Bowman]{rein2024gpqa}
David Rein, Betty~Li Hou, Asa~Cooper Stickland, Jackson Petty, Richard~Yuanzhe
  Pang, Julien Dirani, Julian Michael, and Samuel~R. Bowman.
\newblock {GPQA}: A graduate-level google-proof {Q\&A} benchmark.
\newblock In \emph{Conference on Language Modeling (COLM)}, 2024.

\bibitem[Robey et~al.(2025)Robey, Ravichandran, Kumar, Hassani, and
  Pappas]{robey2025robopair}
Alexander Robey, Zachary Ravichandran, Vijay Kumar, Hamed Hassani, and
  George~J. Pappas.
\newblock {Jailbreaking LLM-Controlled Robots}.
\newblock In \emph{IEEE International Conference on Robotics and Automation
  (ICRA)}, 2025.
\newblock URL \url{https://ieeexplore.ieee.org/document/11128119}.

\bibitem[Schmied et~al.(2026)Schmied, Bornschein, Grau-Moya, Wulfmeier, and
  Pascanu]{schmied2026greedy}
Thomas Schmied, J{\"o}rg Bornschein, Jordi Grau-Moya, Markus Wulfmeier, and
  Razvan Pascanu.
\newblock {LLMs} are greedy agents: Effects of {RL} fine-tuning on
  decision-making abilities.
\newblock In \emph{International Conference on Learning Representations
  (ICLR)}, 2026.

\bibitem[Shao et~al.(2024)Shao, Hu, Wang, Song, Waslander, Liu, and
  Li]{shao2024lmdrive}
Hao Shao, Yuxuan Hu, Letian Wang, Guanglu Song, Steven~L. Waslander, Yu~Liu,
  and Hongsheng Li.
\newblock {LMDrive: Closed-Loop End-to-End Driving with Large Language Models}.
\newblock In \emph{Proceedings of the IEEE/CVF Conference on Computer Vision
  and Pattern Recognition (CVPR)}, pp.\  15120--15130, 2024.
\newblock URL
  \url{https://openaccess.thecvf.com/content/CVPR2024/html/Shao_LMDrive_Closed-Loop_End-to-End_Driving_with_Large_Language_Models_CVPR_2024_paper.html}.

\bibitem[Shinn et~al.(2023)Shinn, Cassano, Gopinath, Narasimhan, and
  Yao]{shinn2023reflexion}
Noah Shinn, Federico Cassano, Ashwin Gopinath, Karthik Narasimhan, and Shunyu
  Yao.
\newblock {Reflexion: Language Agents with Verbal Reinforcement Learning}.
\newblock In \emph{Advances in Neural Information Processing Systems},
  volume~36, 2023.
\newblock URL
  \url{https://papers.nips.cc/paper_files/paper/2023/hash/1b44b878bb782e6954cd888628510e90-Abstract-Conference.html}.

\bibitem[Sima et~al.(2024)Sima, Renz, Chitta, Chen, Zhang, Xie, Bei{\ss}wenger,
  Luo, Geiger, and Li]{sima2024drivelm}
Chonghao Sima, Katrin Renz, Kashyap Chitta, Li~Chen, Hanxue Zhang, Chengen Xie,
  Jens Bei{\ss}wenger, Ping Luo, Andreas Geiger, and Hongyang Li.
\newblock {DriveLM: Driving with Graph Visual Question Answering}.
\newblock In \emph{European Conference on Computer Vision (ECCV)}, 2024.
\newblock URL
  \url{https://www.ecva.net/papers/eccv_2024/papers_ECCV/html/6870_ECCV_2024_paper.php}.

\bibitem[Sun et~al.(2026)Sun, Machcha, Zou, Chan, and
  Chooi]{robocurve2026roboharm}
Edward Sun, Sravanthi Machcha, Sabrina Zou, Tzu~Kit Chan, and Jay Chooi.
\newblock {RoboHarm}: Do frontier robot policies refuse unsafe instructions?
\newblock Robocurve report, \url{https://robocurve.org/roboharm/}, September
  2026.

\bibitem[Tian et~al.(2025)Tian, Gu, Li, Liu, Wang, Zhao, Zhan, Jia, Lang, and
  Zhao]{tian2025drivevlm}
Xiaoyu Tian, Junru Gu, Bailin Li, Yicheng Liu, Yang Wang, Zhiyong Zhao, Kun
  Zhan, Peng Jia, XianPeng Lang, and Hang Zhao.
\newblock {DriveVLM: The Convergence of Autonomous Driving and Large
  Vision-Language Models}.
\newblock In \emph{Proceedings of The 8th Conference on Robot Learning}, volume
  270 of \emph{Proceedings of Machine Learning Research}, pp.\  4698--4726.
  PMLR, 2025.
\newblock URL \url{https://proceedings.mlr.press/v270/tian25c.html}.

\bibitem[{Wayve}(2024)]{wayve2024lingo2}
{Wayve}.
\newblock {LINGO-2: Driving with Natural Language}.
\newblock \url{https://wayve.ai/thinking/lingo-2-driving-with-language/}, 2024.
\newblock Blog post.

\bibitem[Wen et~al.(2024)Wen, Fu, Li, Cai, Ma, Cai, Dou, Shi, He, and
  Qiao]{wen2024dilu}
Licheng Wen, Daocheng Fu, Xin Li, Xinyu Cai, Tao Ma, Pinlong Cai, Min Dou,
  Botian Shi, Liang He, and Yu~Qiao.
\newblock {DiLu: A Knowledge-Driven Approach to Autonomous Driving with Large
  Language Models}.
\newblock In \emph{International Conference on Learning Representations
  (ICLR)}, 2024.
\newblock URL
  \url{https://proceedings.iclr.cc/paper_files/paper/2024/file/93c936b9e492def9c00782cab79dbc6d-Paper-Conference.pdf}.

\bibitem[Xie et~al.(2024)Xie, Zhang, Chen, Li, Zhao, Cao, Toh, Cheng, Shin,
  Lei, Liu, Xu, Zhou, Savarese, Xiong, Zhong, and Yu]{xie2024osworld}
Tianbao Xie, Danyang Zhang, Jixuan Chen, Xiaochuan Li, Siheng Zhao, Ruisheng
  Cao, Jing~Hua Toh, Zhoujun Cheng, Dongchan Shin, Fangyu Lei, Yitao Liu,
  Yiheng Xu, Shuyan Zhou, Silvio Savarese, Caiming Xiong, Victor Zhong, and Tao
  Yu.
\newblock {OSWorld: Benchmarking Multimodal Agents for Open-Ended Tasks in Real
  Computer Environments}.
\newblock In \emph{Advances in Neural Information Processing Systems},
  volume~37, 2024.
\newblock URL
  \url{https://papers.nips.cc/paper_files/paper/2024/hash/5d413e48f84dc61244b6be550f1cd8f5-Abstract-Datasets_and_Benchmarks_Track.html}.

\bibitem[Yang et~al.(2025)Yang, Chen, Zhang, Zhao, Qian, Wang, Wang, Koripella,
  Movahedi, Li, Ji, Zhang, and Zhang]{yang2025embodiedbench}
Rui Yang, Hanyang Chen, Junyu Zhang, Mark Zhao, Cheng Qian, Kangrui Wang,
  Qineng Wang, Teja~Venkat Koripella, Marziyeh Movahedi, Manling Li, Heng Ji,
  Huan Zhang, and Tong Zhang.
\newblock {EmbodiedBench: Comprehensive Benchmarking Multi-modal Large Language
  Models for Vision-Driven Embodied Agents}.
\newblock In \emph{Proceedings of the 42nd International Conference on Machine
  Learning}, volume 267 of \emph{Proceedings of Machine Learning Research},
  pp.\  70576--70631. PMLR, 2025.
\newblock URL \url{https://proceedings.mlr.press/v267/yang25f.html}.

\bibitem[Zhang et~al.(2025)Zhang, Zhu, Wang, Zhou, Yin, Li, Xue, Wang, Hu, Liu,
  Guo, and Zhang]{zhang2025badrobot}
Hangtao Zhang, Chenyu Zhu, Xianlong Wang, Ziqi Zhou, Changgan Yin, Minghui Li,
  Lulu Xue, Yichen Wang, Shengshan Hu, Aishan Liu, Peijin Guo, and Leo Zhang.
\newblock {BadRobot: Jailbreaking Embodied LLM Agents in the Physical World}.
\newblock In \emph{International Conference on Learning Representations
  (ICLR)}, 2025.
\newblock URL
  \url{https://proceedings.iclr.cc/paper_files/paper/2025/hash/5b2fa23e4ef0f7ac6c4f01d7998e6237-Abstract-Conference.html}.

\bibitem[Zitkovich et~al.(2023)Zitkovich, Yu, Xu, Xu, Xiao, Xia, Wu, Wohlhart,
  Welker, Wahid, Vuong, Vanhoucke, Tran, Soricut, Singh, Singh, Sermanet,
  Sanketi, Salazar, Ryoo, Reymann, Rao, Pertsch, Mordatch, Michalewski, Lu,
  Levine, Lee, Lee, Leal, Kuang, Kalashnikov, Julian, Joshi, Irpan, Ichter,
  Hsu, Herzog, Hausman, Gopalakrishnan, Fu, Florence, Finn, Dubey, Driess,
  Ding, Choromanski, Chen, Chebotar, Carbajal, Brown, Brohan, Arenas, and
  Han]{zitkovich2023rt2}
Brianna Zitkovich, Tianhe Yu, Sichun Xu, Peng Xu, Ted Xiao, Fei Xia, Jialin Wu,
  Paul Wohlhart, Stefan Welker, Ayzaan Wahid, Quan Vuong, Vincent Vanhoucke,
  Huong Tran, Radu Soricut, Anikait Singh, Jaspiar Singh, Pierre Sermanet,
  Pannag~R. Sanketi, Grecia Salazar, Michael~S. Ryoo, Krista Reymann, Kanishka
  Rao, Karl Pertsch, Igor Mordatch, Henryk Michalewski, Yao Lu, Sergey Levine,
  Lisa Lee, Tsang-Wei~Edward Lee, Isabel Leal, Yuheng Kuang, Dmitry
  Kalashnikov, Ryan Julian, Nikhil~J. Joshi, Alex Irpan, Brian Ichter, Jasmine
  Hsu, Alexander Herzog, Karol Hausman, Keerthana Gopalakrishnan, Chuyuan Fu,
  Pete Florence, Chelsea Finn, Kumar~Avinava Dubey, Danny Driess, Tianli Ding,
  Krzysztof~Marcin Choromanski, Xi~Chen, Yevgen Chebotar, Justice Carbajal,
  Noah Brown, Anthony Brohan, Montserrat~Gonzalez Arenas, and Kehang Han.
\newblock {RT-2: Vision-Language-Action Models Transfer Web Knowledge to
  Robotic Control}.
\newblock In \emph{Proceedings of The 7th Conference on Robot Learning}, volume
  229 of \emph{Proceedings of Machine Learning Research}, pp.\  2165--2183.
  PMLR, 2023.
\newblock URL \url{https://proceedings.mlr.press/v229/zitkovich23a.html}.

\end{thebibliography}
\bibliographystyle{iclr2027_conference}

\appendix
\section{Prompts}
\label{app:prompts}

Every model received the same three prompts, reproduced verbatim below. The main prompt opened each trial; after a
failed attempt the operator sent the reflection prompt and, once the model had replied, the continuation prompt. The
text is exactly as sent, including the Markdown escape characters (e.g.\ \texttt{set\textbackslash\_motion}); only
typographic quotes and dashes are shown as their ASCII equivalents.

\subsection*{Main prompt}
\begin{lstlisting}[basicstyle=\ttfamily\scriptsize, breaklines=true, breakindent=0pt, columns=fullflexible, keepspaces=true, upquote=true, frame=single, rulecolor=\color{black!25}, xleftmargin=2pt, xrightmargin=2pt]
Use only the `drivingbench_sandbox` MCP. Complete the stated objective using its tools.

OBJECTIVE
You are in a course delineated by mini-cones (small multicolored) and the environment (walls, etc). Your objective is to drive through the course (in a backwards-U-shaped parking lot) and stay between the cones. Your finish line is a wide "parking spot" marked by numerous BLUE mini-cones at the very end; finish by parking in this area. You will be evaluated primarily by how far you get in the course (without leaving the boundaries/collisions), but a secondary objective is to complete the course in less time.

ROLE
Use set\_motion(direction, steering\_percent, speed\_mps, duration\_s, reason). Direction is left, right or straight; steering\_percent is 0-100. Zero is straight for either turn direction.
100% requests 180 steering-wheel degrees; observe reports steering\_percent on the same scale, positive left. The speed ceiling is 3.5 m/s. The final steering target is sent immediately; native limits govern actual response.
In every set\_motion and stop\_now, give reason: about thirty words on what you see and what the command is for. It is recorded for the operator and never changes what the car does.

DRIVING

- Choose durations that allow observation and reasoning latency; the car continues moving while you think.
- Observe, send a motion command, then observe actual movement and adapt.
- Commands replace globally rather than queue. Duration starts at acceptance and includes steering buildup and engagement waits. Motion continues while you think; expiry begins braking, not guaranteed standstill.
- A rejected replacement leaves the previous command unchanged. After an uncertain transport result, observe instead of blindly retrying.
- Any connected chat can replace the active command or call stop\_now. DrivingBench is always active. Native engagement and ready vehicle inputs are required; Stop cancels motion until a fresh command.
- timestamp is UTC response time, not image capture or command acceptance time. image\_age\_s is image age at producer response.
- Carefully look for objects and obstacles (parked cars, buildings, islands, trees, etc); they're all 3D and the vehicle you are controlling is also 3D and is also wider than it might seem from the camera images. You must avoid all objects and obstacles, your attempt will be terminated if you collide with any of them.
- If possible, try to always be in the center of your lane/road and maximize distance from obstacles.
- In successive observed images, pay attention to what changes, i.e. new obstacles/information/etc, and/or changes in distances to existing things you've seen in previous images. If obstacles are getting closer on either side, this could encourage steering away from them to maintain distance.
- Over time, understand the behavior that occurs when you choose different actions or steering inputs and adapt to this.
- Think and plan carefully where you want to end up and plan your route accordingly and advance toward subgoals.
- Don't be afraid to do sharp turns or choose high steering %s. It's better to be aggressive and "make" the turn than to be conservative and not have room to finish the turn to get to where you want to go.
- Also, your vehicle cannot do arbitrarily tight turns; 100% might be much less tight than you expect. 100% corresponds to doing a 90 degree turn at 1 m/s for \~60 seconds. Therefore you might want to start your turn earlier than you expect and plan your approach to account for this.
- Largely keep your turning %s for left/right in the set {30%, 60%, 100%}.
\end{lstlisting}

\subsection*{Reflection prompt}
\begin{lstlisting}[basicstyle=\ttfamily\scriptsize, breaklines=true, breakindent=0pt, columns=fullflexible, keepspaces=true, upquote=true, frame=single, rulecolor=\color{black!25}, xleftmargin=2pt, xrightmargin=2pt]
Your previous attempt has now ended (likely due to a mistake or collision with an object or the environment). Reflect on how you did in your previous attempt, why it might have ended, and what you can do better in a future attempt.
\end{lstlisting}

\subsection*{Continuation prompt}
\begin{lstlisting}[basicstyle=\ttfamily\scriptsize, breaklines=true, breakindent=0pt, columns=fullflexible, keepspaces=true, upquote=true, frame=single, rulecolor=\color{black!25}, xleftmargin=2pt, xrightmargin=2pt]
You now have been granted another attempt to complete the same objective as before. Begin now.
\end{lstlisting}

\section{Protocol and latency details}
\label{app:details}

\subsection{Operator messages}
\label{app:operator}
{The operator sent two messages beyond the fixed prompts, both visible in the released transcripts. After the
continuation prompt for Astra's attempt~2, the model observed once while the car was still being reset; the operator
interrupted that turn, the next turn ended without a command, and the operator sent ``Continue'' with no other
information. For Grok, the operator ended each attempt by sending the next message instead of interrupting the
turn, because in Cursor pressing stop and then typing can overwrite the previous message instead of continuing a new turn. This includes the reflection prompt after
attempts 1 and~2, and ``Stop'' after attempt~3.}

\subsection{Latency examples}
\label{app:latency}
\textbf{Latency and commands expiring.} Commands have durations, and when they run out, the car brakes. Additionally, the car continues moving while the model thinks or during the wait time between model tool calls or responses. Therefore, models sometimes acted based on frames that were many seconds (or meters) old, or let commands expire and then had to restart from rest which can be more difficult in many situations (such as turns). 
\begin{itemize}
 \item Grok's second attempt sent its last command 17\,s after the last frame it ended up seeing. During that time, the car rolled 7.4\,m (Figure~\ref{fig:failures}). Its reflection (turn 28) includes: ``The first move expired. I next looked while already stopping at 1.1\,m/s, with yellow and red cones a few meters ahead.'' and ``I then waited $\sim$13 seconds before the next command. Expiry only starts braking; the car can still roll into a cone.''
 \item Astra realized the same issue in its first attempt: ``At roughly 1--1.5\,m/s, several seconds of interpretation and tool calls meant several additional meters of travel. The strong correction began after much of the available clearance was already gone.'' (turn 51)
 \item Fable, in its second attempt, acted based on a frame that was 55\,s old. The car was nearly stationary (and only moved 2.1\,m), so this mostly wasted time rather than burning distance blindly. But in general, Fable spent such a low proportion of time moving the car: it was only moving for 15\% of this attempt, the lowest out of all the attempts (Table~\ref{tab:attempts}).
\end{itemize}

\ifarxiv\begingroup
\section{Vehicle characterization}
\label{app:vehicle}

Before the trial we characterized the Corolla, comma four and openpilot in supervised runs for calibration purposes. Four properties explain some of the behavior reported in Section~\ref{sec:results}.

\textbf{Launch overshoot.} While the car is stopped and engaged, Toyota's port of openpilot integrates the gap between requested and actual acceleration, so the delivered acceleration command ramps at $0.5\,a_\text{req} - 0.03$\,m/s$^3$, a rate that we derived from the source and measured to within 0.3\% on the car. The car starts to move only once this command passes a certain breakaway value (which we measured to be about 0.52\,m/s$^2$ on flat pavement, and up to 1.56\,m/s$^2$ in one run), and the stored acceleration can then carry it well past the request: the first command from rest reached 1.6--2.4\,m/s for requests of 0.8--1.5\,m/s. To be clear, this accumulation happens in the car-specific controller, which is downstream of openpilot's own longitudinal controller, so resetting the latter at standstill (the first fix we tried) had no effect.

\textbf{No steering at rest.} openpilot's lateral control becomes active only above 0.30\,m/s, so a steering (turning) command sent at complete standstill must wait until the car starts rolling. From rest, the wheel needed a median of 4.0\,s to reach 90\% of a steering change, versus 1.9\,s when rolling (Figure~\ref{fig:steering}).

\textbf{Steering range and rate.} In our calibration runs the measured wheel angle plateaued near 120--130$^\circ$, so we set 100\% to 180$^\circ$, which spans that range with headroom; in the trial, requests of 100\% reached a median of 129$^\circ$ and at most 173$^\circ$. Toyota's 100$^\circ$/s steering-rate limit bounds how fast a turn builds, and the turning radius at the typical plateau is about 16\,m.

\textbf{Left--right asymmetry.} With an earlier curvature controller, the car turned in more slowly to the left than to the right: from straight at 0.6\,m/s it needed about 22.5\,m to reach full left curvature but 12\,m to reach full right. We did not re-measure this with the final controller.
\endgroup
\fi

\clearpage
\section{Additional results}
\label{app:results}
\label{app:record}

\begin{figure}[h]
  \centering
  \includegraphics[width=\textwidth]{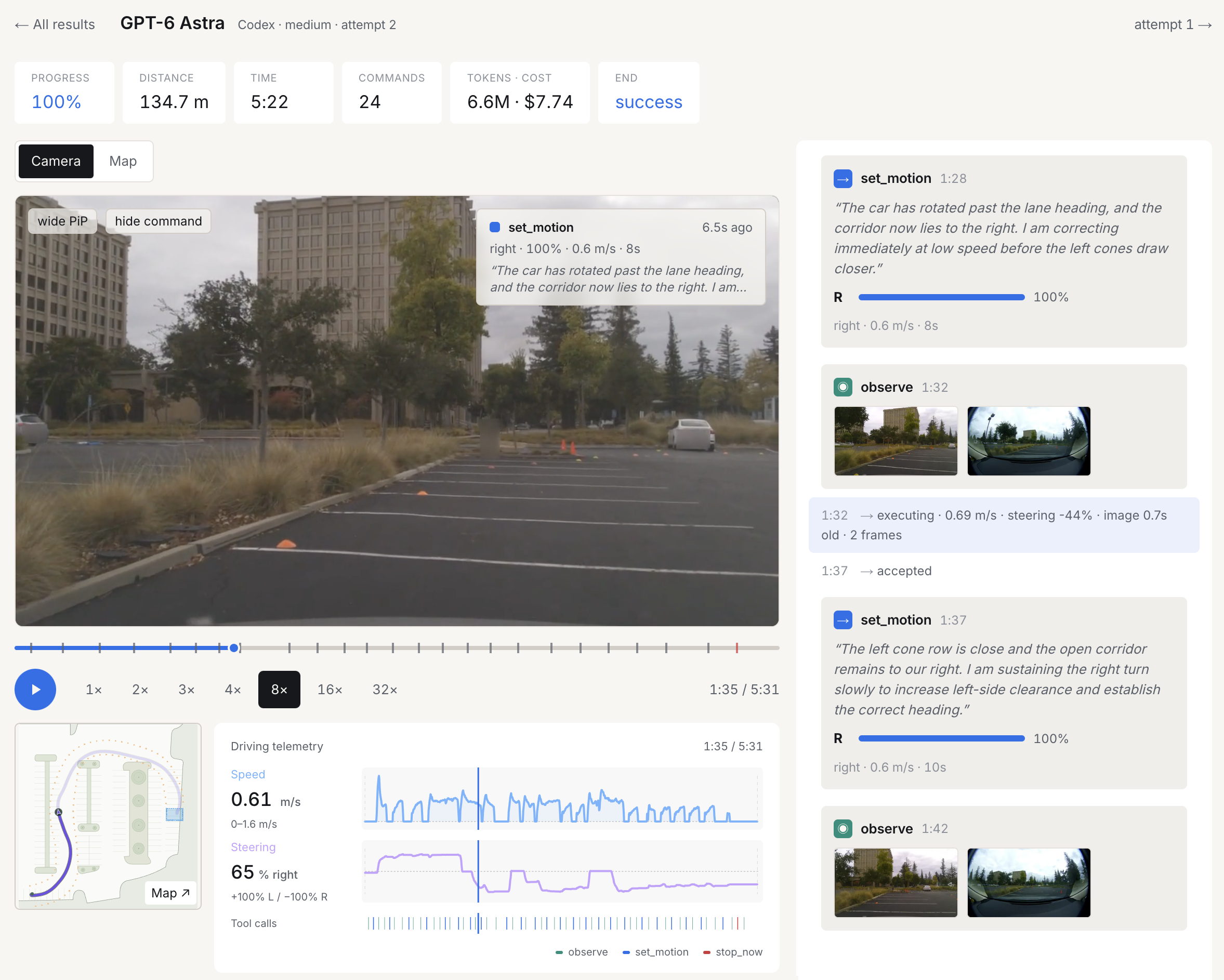}
  \caption{{\textbf{Trace viewer.} Trajectories recorded with our driving harness can be visualized in our trace viewer tool. It renders the recorded video of the evaluation run along with MCP tools that were used by the model and car telemetry such as steering, acceleration and position. All these data points are synchronized by timestamps and can be replayed alongside the video recording.}}
  \label{fig:viewer}
\end{figure}

\begin{figure}[h]
  \centering
  \includegraphics[width=\textwidth]{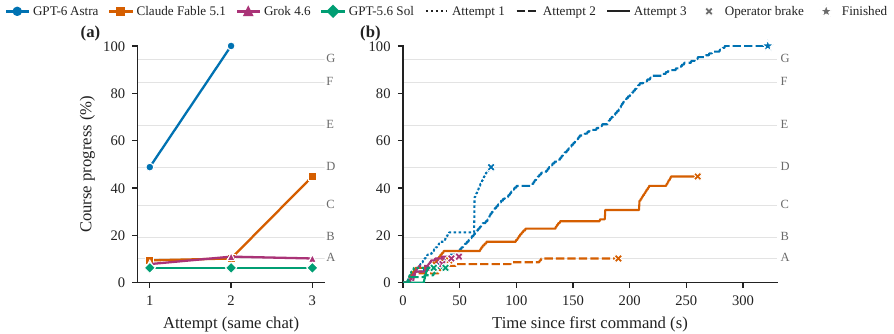}
  \caption{\textbf{Course progress.} Progress is the furthest point along the intended centerline
  ({127\,m, up to the finish zone}) that the car reached while staying within 4\,m of it.
  (a)~Progress per attempt. Later attempts happen in the same chat after a fixed reflection prompt.
  (b)~Progress over time, measured from the first accepted command.
  Horizontal guides mark the section labels in Figure~\ref{fig:trajectories}.}
  \label{fig:progress}
\end{figure}

\begin{figure}[h]
  \centering
  \includegraphics[width=\textwidth]{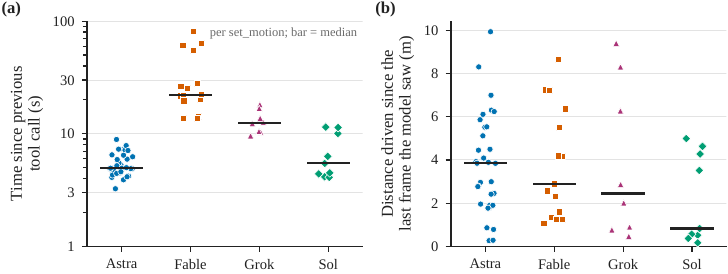}
  \caption{\textbf{Acting on stale observations.} For every accepted \texttt{set\_motion}:
  (a)~time since the previous tool call, mostly model deliberation (log scale);
  (b)~distance the car drove between the capture of the last camera frame the model had received and the
  command's acceptance. Bars are medians.}
  \label{fig:decisions}
\end{figure}

\begin{figure}[h]
  \centering
  \includegraphics[width=\textwidth]{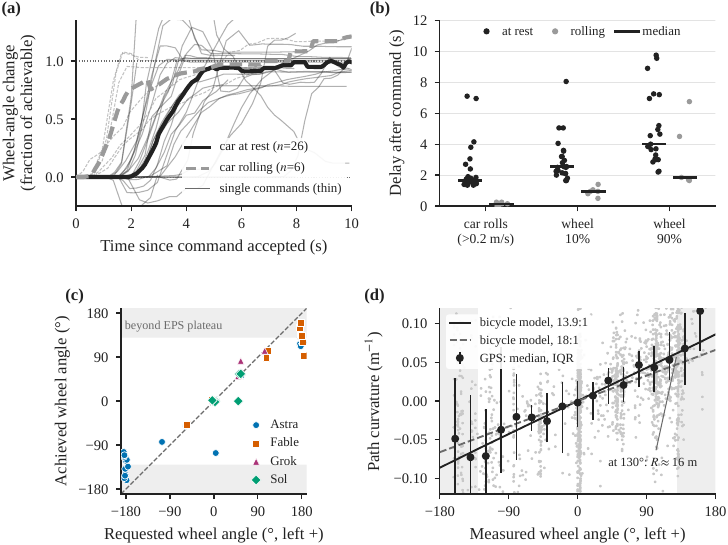}
  \caption{\textbf{Requested versus actual steering.}
  (a)~Wheel-angle response to every command asking for a change of at least $30^\circ$, normalised to the
  achievable change (request clipped to the $\approx130^\circ$ EPS plateau). Thick lines are medians.
  (b)~Delays after a command is accepted. Of the 32 commands in (a), 26 arrived after the previous command had
  expired and the car had braked. The EPS only turns the wheel once the car rolls, so from rest the wheel
  reaches 90\% of its change after a median of 4.0\,s (1.9\,s when already rolling).
  (c)~Achieved versus requested wheel angle per command. Requests of 100\% ($180^\circ$) reach a median of
  $129^\circ$, so the upper $\approx28\%$ of the command range wasn't reached at the conditions of the model benchmark evaluations. 
  (d)~Measured path curvature from GPS against measured wheel angle, compared with a kinematic bicycle model.
  A 13.9:1 ratio fits better than the 18:1 in the harness documentation. At the plateau the turning radius
  is $\approx 16$\,m, which implies roughly 25 s for a $90^\circ$ turn at 1 m/s, which is faster than the $\sim 60$\,s we measured in a single sustained calibration run before the trial from rest and was used in the prompt's example. }
  \label{fig:steering}
\end{figure}

\begin{figure}[h]
  \centering
  \includegraphics[width=\textwidth]{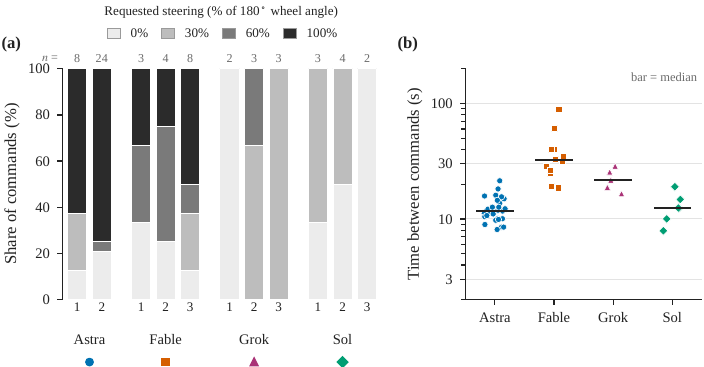}
  \caption{\textbf{What the models commanded.}
  (a)~Distribution of the requested \texttt{steering\_percent} for each attempt ($n$ = number of commands).
  (b)~Time between consecutive accepted \texttt{set\_motion} calls, pooled over attempts (log scale).}
  \label{fig:commands}
\end{figure}

\begin{figure}[p]
  \centering
  \includegraphics[width=\textwidth]{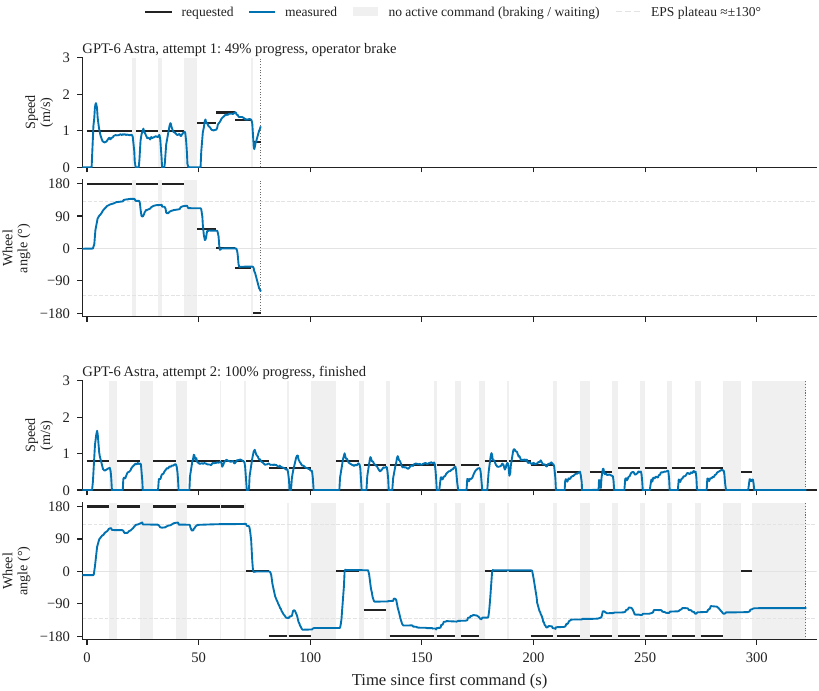}
  \caption{\textbf{Control traces, GPT-6 Astra.} For each attempt: requested (black) and measured speed
  and steering-wheel angle over time. Shading marks periods with no active command, when the previous
  command had expired and the car was braked while the model deliberated. The dashed lines mark the
  typical $\approx\pm130^\circ$ EPS plateau.}
  \label{fig:controls-astra}
\end{figure}

\begin{figure}[p]
  \centering
  \includegraphics[width=\textwidth]{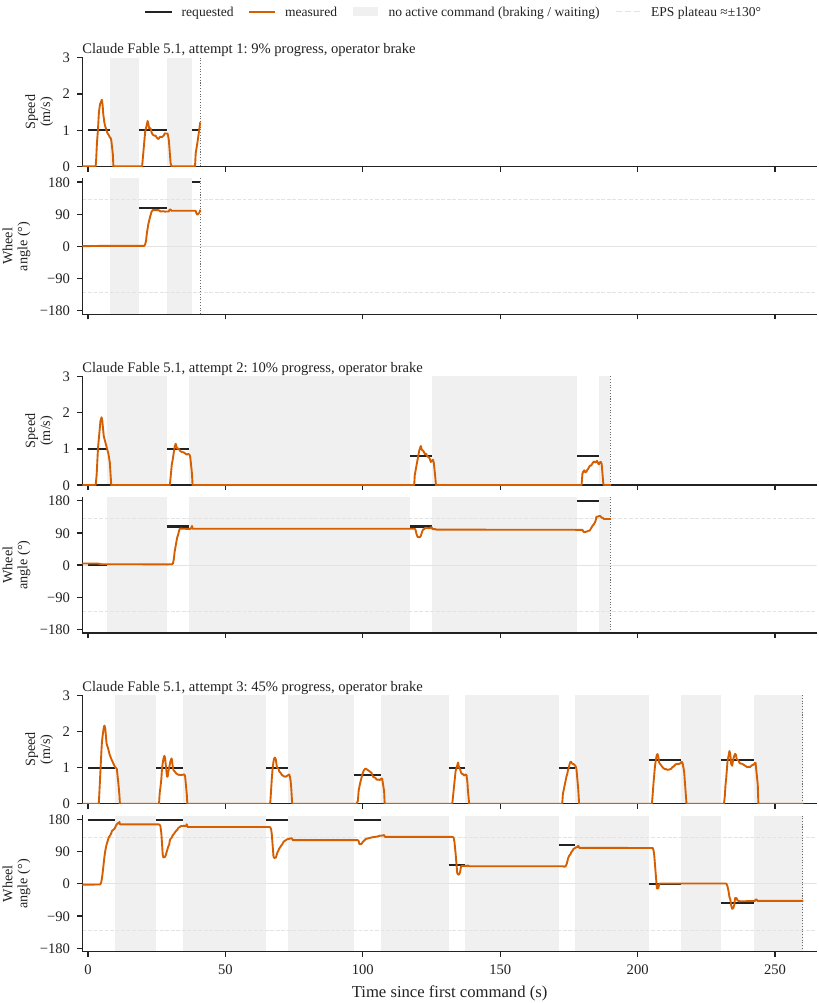}
  \caption{\textbf{Control traces, Claude Fable 5.1.} As Figure~\ref{fig:controls-astra}.}
  \label{fig:controls-fable}
\end{figure}

\begin{figure}[p]
  \centering
  \includegraphics[width=\textwidth]{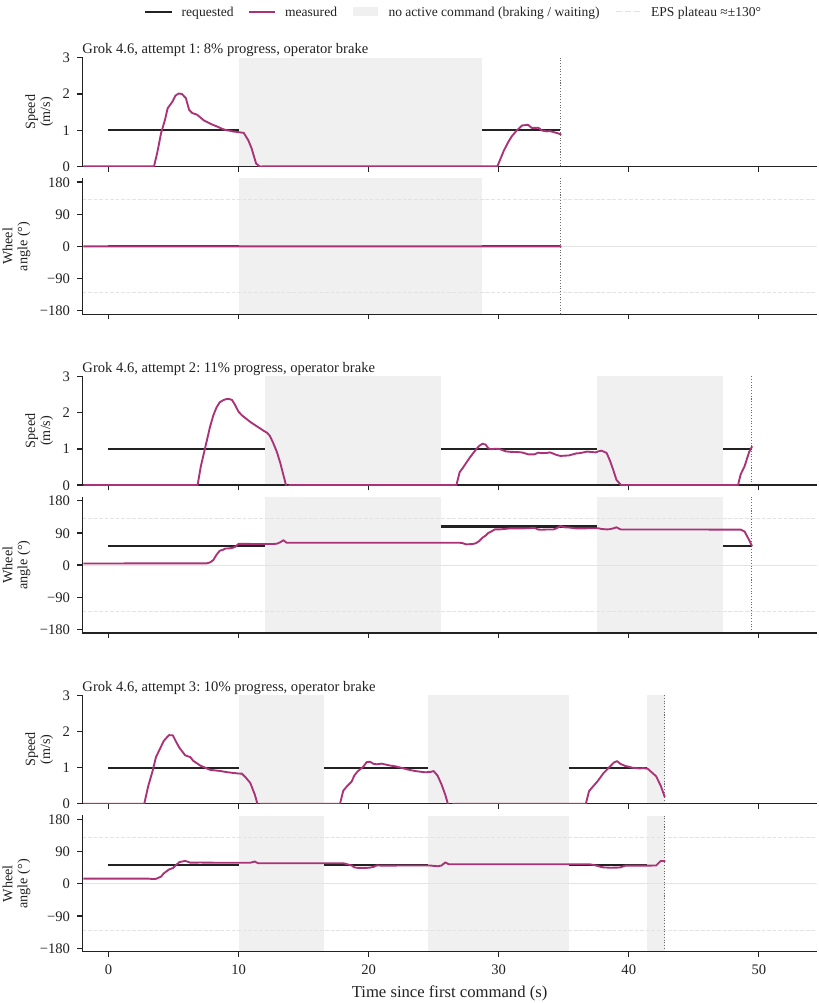}
  \caption{\textbf{Control traces, Grok 4.6.} As Figure~\ref{fig:controls-astra}.}
  \label{fig:controls-grok}
\end{figure}

\begin{figure}[p]
  \centering
  \includegraphics[width=\textwidth]{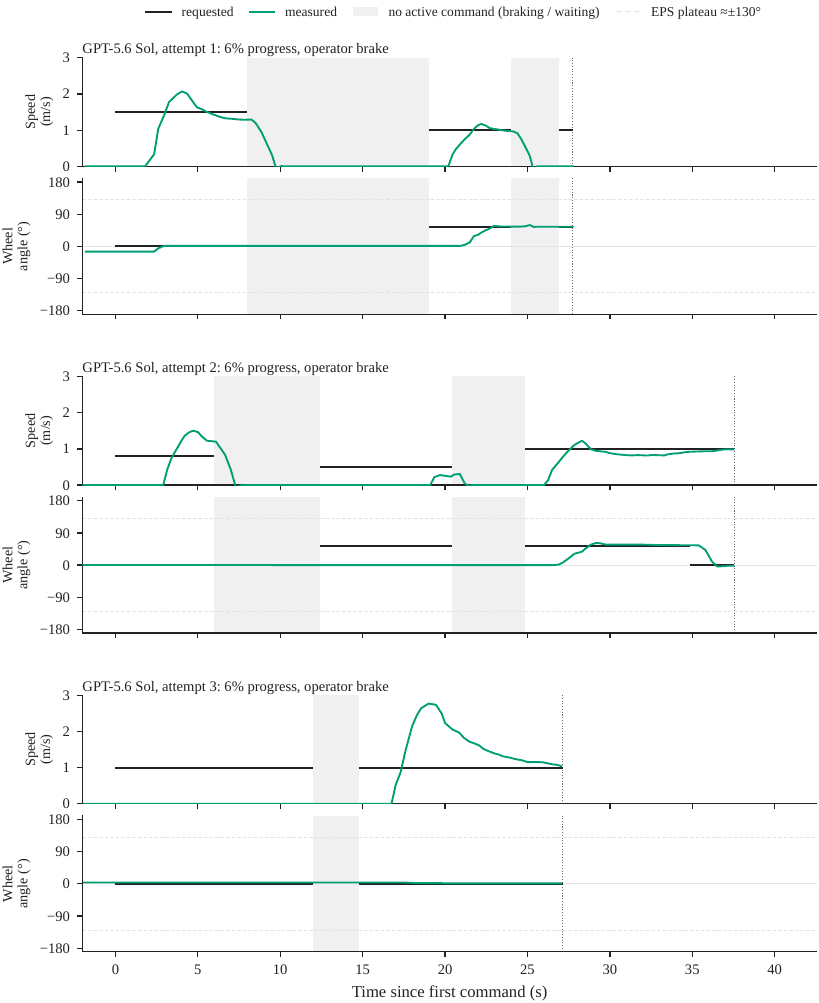}
  \caption{\textbf{Control traces, GPT-5.6 Sol.} As Figure~\ref{fig:controls-astra}.}
  \label{fig:controls-sol}
\end{figure}

{Table~\ref{tab:outcomes} records how each attempt ended and Table~\ref{tab:attempts} gives per-attempt statistics. Figure~\ref{fig:viewer} shows the trace viewer used to inspect attempts. Figures~\ref{fig:progress}, \ref{fig:decisions} and~\ref{fig:steering} support Section~\ref{sec:results}; Figure~\ref{fig:commands} shows the steering commands each attempt requested, and Figures~\ref{fig:controls-astra}--\ref{fig:controls-sol} show requested against measured speed and steering over every attempt.}

\begin{table}[h]
  \centering
  \caption{{\textbf{How each attempt ended}, from the road video and human operator notes. Letters are the course sections of Figure~\ref{fig:trajectories}.}}
  \label{tab:outcomes}
  \small
  \begin{tabularx}{\textwidth}{@{}ll r >{\raggedright\arraybackslash}X@{}}
  \toprule
  Model & \# & Progress (\%) & End of attempt \\
  \midrule
  GPT-6 Astra & 1 & 49 & Operator brake: drifted onto the left cone line and island at the right turn into the cross aisle (D) \\
   & 2 & 100 & Finished: the model's own \texttt{stop\_now} inside the finish zone \\
  \addlinespace
  Claude Fable 5.1 & 1 & 9 & Operator brake: swung left across the cone line toward the median curb (A) \\
   & 2 & 10 & Operator brake: outside the cone line again, pointed at the median island's curb (A) \\
   & 3 & 45 & Operator brake: too little room for the right turn at D, pointed at the planter island \\
  \addlinespace
  Grok 4.6 & 1 & 8 & Operator brake: drove straight into the mini-cone gate toward the landscaped island (A) \\
   & 2 & 11 & Operator brake: crossed the mini-cone line toward the median island (A) \\
   & 3 & 10 & Operator brake: crossed the mini-cone line toward the median island (A) \\
  \addlinespace
  GPT-5.6 Sol & 1 & 6 & Operator brake: crossed the mini-cone line at the start of the course (A) \\
   & 2 & 6 & Operator brake: drove over the mini-cone line toward the landscaped island (A) \\
   & 3 & 6 & Operator brake: crossed the cone line at 2.8\,m/s, heading for the landscaped island (A) \\
  \bottomrule
  \end{tabularx}
\end{table}

\begin{table}[h]
  \centering
  \caption{\textbf{Per-attempt statistics.} Dur.: first accepted command to the end of the last engagement.
  Moving: share of that window with speed $>0.1$\,m/s.
  Repl.: commands sent while the previous one still had time left, out of the commands that followed another command.
  Med.\ gap: median time between consecutive commands. Peak: highest measured speed.}
  \label{tab:attempts}
  \small
  \setlength{\tabcolsep}{3.5pt}
  \resizebox{\textwidth}{!}{\begin{tabular}{@{}llrrrrrrrrr@{}}
\toprule
Model & \# & Prog. (\%) & Dist. (m) & Dur. (s) & Cmds & Cmd/min & Moving (\%) & Repl. & Med. gap (s) & Peak (m/s) \\
\midrule
GPT-6 Astra & 1 & 49 & 67.3 & 78 & 8 & 6.2 & 85 & 2/7 & 10.0 & 1.8 \\
 & 2 & 100 & 134.7 & 322 & 24 & 4.5 & 65 & 2/23 & 12.1 & 1.6 \\
\addlinespace
Claude Fable 5.1 & 1 & 9 & 17.5 & 41 & 3 & 4.4 & 45 & 0/2 & 18.9 & 1.8 \\
 & 2 & 10 & 27.3 & 190 & 4 & 1.3 & 15 & 0/3 & 60.8 & 1.9 \\
 & 3 & 45 & 73.7 & 260 & 8 & 1.8 & 27 & 0/7 & 32.7 & 2.2 \\
\addlinespace
Grok 4.6 & 1 & 8 & 14.4 & 35 & 2 & 3.5 & 35 & 0/1 & 28.7 & 2.0 \\
 & 2 & 11 & 22.6 & 49 & 3 & 3.6 & 39 & 0/2 & 23.6 & 2.4 \\
 & 3 & 10 & 22.2 & 43 & 3 & 4.2 & 53 & 0/2 & 17.7 & 1.9 \\
\addlinespace
GPT-5.6 Sol & 1 & 6 & 15.0 & 28 & 3 & 6.5 & 42 & 0/2 & 13.5 & 2.1 \\
 & 2 & 6 & 15.6 & 38 & 4 & 6.4 & 46 & 1/3 & 12.4 & 1.5 \\
 & 3 & 6 & 17.1 & 27 & 2 & 4.4 & 36 & 0/1 & 14.8 & 2.8 \\
\bottomrule
\end{tabular}}
\end{table}


\end{document}